\pdfoutput=1

\documentclass[runningheads]{llncs}
\RequirePackage{silence}  
\usepackage{graphicx}
\usepackage{comment}
\usepackage{amsmath,amssymb}
\usepackage{color}
\usepackage{url}
\usepackage{hyperref}
\usepackage{microtype}
\hypersetup{
	pdftitle={Mask What Matters: Saliency-Guided Video Self-Supervised Learning for Autonomous Driving},
	pdfauthor={Christopher Lang, Alexander Braun, Abhinav Valada}
}

\usepackage[acronym]{glossaries}
\newacronym{mae}{MAE}{masked auto-encoding}
\newacronym{vla}{VLA}{vision-language action model}
\newacronym{ssl}{SSL}{self-supervised learning}
\newacronym{vlm}{VLM}{vision-language model}
\newacronym{dpt}{DPT}{dense prediction transformer}
\newacronym{vit}{ViT}{vision transformer}
\newacronym{mcd}{MCD}{motion-corrected difference}
\newacronym{ema}{EMA}{exponential moving average}
\newacronym{mot}{MOT}{multi-object tracking}
\newacronym{iou}{IoU}{intersection-over union}
\glsdisablehyper
\usepackage{amsfonts}
\usepackage{booktabs}
\usepackage{multirow}
\usepackage{float}

\usepackage[inline]{enumitem}
\usepackage[table,dvipsnames]{xcolor}
\usepackage{subcaption}
\usepackage{tcolorbox}
\usepackage{siunitx}

\usepackage[capitalize]{cleveref}
\crefname{section}{Sec.}{Secs.}
\Crefname{section}{Section}{Sections}
\Crefname{table}{Table}{Tables}
\crefname{table}{Tab.}{Tabs.}
\crefname{algorithm}{Algo.}{Algos.}

\newcommand{\figref}[1]{Fig.~\ref{#1}}
\newcommand{\tabref}[1]{Tab.~\ref{#1}}
\newcommand{\secref}[1]{Sec.~\ref{#1}}

\newcommand{\publicationfootnote}{%
  \begingroup
  \renewcommand\thefootnote{}%
  \footnotetext{\footnotesize
    To appear in Proceedings of the 48th DAGM German Conference on Pattern Recognition (GCPR), 2026. The final publication will be available through Springer.}%
  \endgroup
}

\begin{document}


\title{Mask What Matters: Saliency-Guided Video Self-Supervised Learning for Autonomous Driving}

\author{Christopher Lang\inst{1,2} \and
Alexander Braun\inst{1} \and
Abhinav Valada\inst{2}}

\authorrunning{C. Lang et al.}

\institute{Robert Bosch GmbH, Stuttgart, Germany \and
Robot Learning Lab, University of Freiburg, Freiburg, Germany\\
\email{\{lang,valada\}@cs.uni-freiburg.de}}

\maketitle  

\publicationfootnote

\begin{abstract}
Video self-supervised learning through masked spatiotemporal prediction has emerged as a promising paradigm for learning feature representations from unlabeled data. 
However, existing methods typically rely on random masking, which indiscriminately removes regions irrespective of their semantic or temporal relevance. 
In ego-centric driving videos, this can weaken the pretext signal since safety-critical cues such as pedestrians, vehicles, lane boundaries, and dynamic interactions often occupy only a small portion of the frame, yet are central to downstream perception. 
We introduce \texttt{V-JEPA4A}, a domain-specialized variant of V-JEPA for autonomous driving that is pre-trained on publicly available driving videos with a novel saliency-driven masking policy. It accounts for semantically and temporally relevant context. 
The proposed policy preserves and predicts context according to semantic importance and temporal relevance, yielding more informative representation learning while retaining the efficiency of masked prediction. 
We evaluate the resulting encoders on four driving benchmarks spanning tracking, semantic segmentation, and depth estimation. 
The results demonstrate that \texttt{V-JEPA4A} reduces identity switches on BDD100k MOT by $25\%$ over V-JEPA with random masking, achieves $73.2$ mIoU on Cityscapes, and $3.75$ RMSE on KITTI-2015 depth, while incurring only ${\sim}14\%$ additional pre-training iteration overhead.

\keywords{Video Self-Supervised Learning \and Masked-Latent Prediction \and Autonomous Driving.}
\end{abstract}

\section{Introduction}
\label{sec:intro}
Perception encoders used in autonomous vehicles must capture fine-grained object geometry, multi-frame temporal dynamics, and scene-level semantic context under continuously changing conditions, while remaining efficient enough for downstream deployment. 
Self-supervised learning (\acrshort{ssl})~\cite{dinov2,clip} has become a powerful paradigm for learning such representations from unlabeled data. 
In particular, recent video \acrshort{ssl} methods based on masked spatiotemporal prediction, either in pixel space~\cite{videomae} or latent space~\cite{ijepa,vjepa}, have shown promising results on multiple general-purpose benchmarks. 
These methods typically rely on random masking, which provides a simple and scalable pretext task. 
However, when transferring such models to traffic scenes, we identify a structural mismatch between content-agnostic random masking strategies applied in these methods and the distinctive characteristics of ego-centric driving videos.

\begin{figure}[t]
    \centering
    \includegraphics[width=0.8\linewidth]{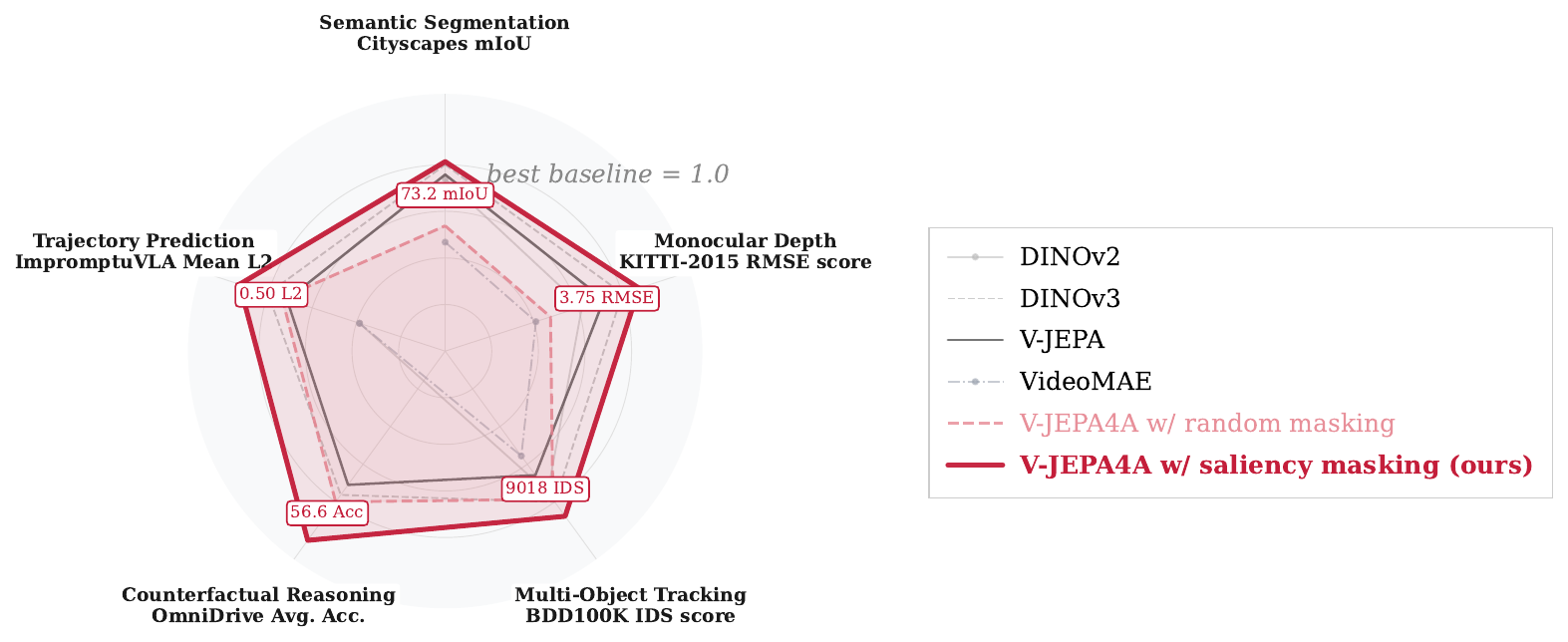}
    \vspace{-0.5em}
    \caption{Radar chart comparing \texttt{V-JEPA4A} with image- and video-based self-supervised baselines across five driving-relevant downstream evaluations.
    All axes report normalized scores relative to the best baseline, where the dashed reference ring denotes baseline parity. 
    For lower-is-better metrics such as RMSE, IDS, and trajectory L2 error, scores are inverted before normalization, so larger values consistently indicate better performance.
    }
    \label{fig:spider_plot}
    \vspace{-0.3cm}
\end{figure}

Driving scenes are dominated by large background regions such as road surfaces, buildings, sky, and parked vehicles, while safety-critical cues often occupy only a small fraction of the image. A pedestrian stepping off the curb, a traffic light changing state, a cyclist preparing to merge, or a distant vehicle initiating a lane change may be spatially small but semantically decisive. Random masking used in existing video \acrshort{ssl} methods~\cite{videomae,vjepa} treats these regions no differently from static background, and can therefore indiscriminately discard the cues that are most informative for learning driving-relevant representations. When such regions are masked without sufficient visible context, the prediction target becomes ambiguous rather than semantically grounded. As a result, the model may be encouraged to reconstruct the dominant scene appearance rather than learn features that discriminate dynamic agents, object boundaries, and temporally salient events. In our experiments, this manifests as weak instance discriminability in multi-object tracking and reduced sensitivity to small road users such as pedestrians and cyclists.

Motivated by this observation, we study the hypothesis that \emph{saliency-driven spatiotemporal masking improves the quality of self-supervised video representations for autonomous driving by encouraging the pretext task to focus on semantically and temporally informative regions.} We instantiate this idea in \texttt{V-JEPA4A}, a driving-specific variant of V-JEPA that modifies only the masking policy while keeping the architecture, prediction objective, and training pipeline otherwise unchanged. This design isolates the effect of masking and allows us to evaluate whether lightweight domain knowledge, injected at the level of the pretext task, can improve downstream perception without additional supervision, losses, or task-specific pre-training.

We investigate saliency signals from three complementary sources: Sobel gradients, optical flow magnitude, and a motion-compensated frame differencing (\acrshort{mcd}). These signals are converted into per-pixel saliency maps and used to guide the selection of foreground and background patches during masking. The resulting policy increases the likelihood that the model observes and predicts semantically and temporally coherent content, rather than uniformly sampling from the image. We pre-train \texttt{V-JEPA4A} from scratch using only publicly available driving videos, with orders of magnitude less data than large-scale general-purpose models such as DINOv2, OpenCLIP, and V-JEPA. The learned encoder is then frozen and evaluated across four driving-relevant downstream tasks, including depth, segmentation, multi-object tracking, and visual question answering.

Our main contributions are as follows:
\begin{enumerate}[topsep=0pt]
    \item We identify and empirically analyze the mismatch between random masking in video \acrshort{ssl} and the spatially sparse, safety-critical structure of egocentric driving scenes.
    \item We present \texttt{V-JEPA4A}, a driving-specific instantiation of V-JEPA that changes only the masking policy, requires no additional supervision or architectural modifications, is pre-trained from scratch on publicly available driving data, and uses orders of magnitude less data than general-purpose alternatives.
    \item We systematically compare four saliency-driven masking strategies based on 
        \begin{enumerate*}
            \item Sobel gradients;
            \item Optical flow magnitude;
            \item Motion-compensated frame differencing;
            \item Combined Sobel\,+\,optical flow;
        \end{enumerate*} 
     and evaluate their impact across four autonomous driving benchmarks.
    \item We demonstrate state-of-the-art performance across three complementary evaluation axes: multi-object tracking on BDD100k (reducing IDS by 25\% over the original V-JEPA), semantic segmentation on Cityscapes (73.2 mIoU), and monocular depth estimation on KITTI-2015 (3.75 RMSE); a consolidated overview is given in \figref{fig:spider_plot}.\looseness=-1
\end{enumerate}

\section{Related Work}
\label{sec:related_work}
In this section, we briefly review the main image-based \acrshort{ssl} paradigms, their extension to video, and adaptations to the driving domain.

{\parskip=2pt
\noindent\textit{Image-Based Self-Supervised Learning}: Image-based \acrshort{ssl} has largely developed along three directions.
\emph{Contrastive methods}~\cite{simclr,moco} learn augmentation-invariant embeddings, an idea that is further extended to image-text alignment by CLIP~\cite{clip} and SigLIP~\cite{siglip}.
\emph{Self-distillation methods} train a student to match the feature representation of a momentum-updated teacher on different views of the same image, without explicit negative pairs. 
DINOv2~\cite{dinov2} scaled this paradigm and incorporated masked-prediction ideas~\cite{ibot}, and DINOv3~\cite{dinov3} further improved dense feature quality during long training schedules, leading to particularly strong performance on dense prediction tasks.
\emph{Masked and predictive methods} reconstruct withheld content as a pretext. 
MAE~\cite{mae} showed that heavily masked-pixel reconstruction yields strong, transferable features.
More recently, latent prediction methods such as I-JEPA~\cite{ijepa} proposed predicting latent targets for masked regions, shifting supervision away from pixel-wise reconstruction toward higher-level semantic structure.}

{\parskip=2pt
\noindent\textit{Self-Supervision for Video}: 
Extending \acrshort{ssl} from images to video requires learning temporal structure in addition to spatial semantics. 
\emph{Contrastive video methods} adapt image-style learning to clips, though some temporal augmentations risk changing clip semantics~\cite{videomoco,cvrl}. 
Recent video foundation models have scaled video-language pre-training substantially by combining masked video modeling, cross-modal contrastive learning, or next-token prediction~\cite{internvideo2,videoclip}.
\emph{Masked video modeling} has become one of the dominant paradigms for video \acrshort{ssl}. 
VideoMAE~\cite{videomae} reconstructs pixels under heavy spatiotemporal masking. 
For the sparse task of action recognition, AdaMAE~\cite{adamae} proposed using a learned masking approach as an alternative to random multi-block masking, providing early indications that informed masking is beneficial for learning global feature representations. 
MGMAE~\cite{mgmae} further explored propagating the mask blocks based on optical flow to prevent information leakage. 
Analogous to the pixel-level prediction, V-JEPA~\cite{vjepa} applied the prediction of masked spatiotemporal features in latent space to the video domain. 
In contrast to JEPA's \acrshort{ema}-updated target encoder, SALT~\cite{salt} showed that the teacher can also be a frozen encoder pre-trained on a different task, greatly reducing computational requirements.}

{\parskip=2pt
\noindent\textit{Autonomous Driving}: 
While the above approaches are effective in general settings, they do not account for domain-specific structure.
Autonomous driving places additional demands on representation learning: scenes are dominated by continuous ego-motion, and rare but safety-relevant events must be represented. 
Driving-pretraining specific work focuses mainly on voxel feature learning from multi-view inputs~\cite{visionpad}, which limits the available scenes for pre-training to those captured by multi-view sensor setups. 
Other driving-based end-to-end models still rely on image-based pre-training~\cite{omnidrive,impromptuvla}. 
DINOv2~\cite{dinov2}, DINOv3~\cite{dinov3}, OpenCLIP~\cite{clip}, and V-JEPA~\cite{vjepa} are designed as general-purpose representation backbones, trained on internet-scale corpora with content-agnostic objectives and intended to transfer broadly across tasks.
In perception domains such as autonomous driving, however, the relevant statistics, geometry, and safety-critical cues differ substantially from the object-centric distributions these models were optimized for, and re-training at a general-purpose scale is rarely an option due to data, compute, and license constraints.
Our work positions \texttt{V-JEPA4A} as a \emph{domain-specific} counterpart to these models: it inherits the latent masked-prediction training objective but is pre-trained with a masking policy tailored to driving tasks.}

\section{Technical Approach}
\label{sec:system_overview}
\begin{figure*}[t]
    \centering
    \includegraphics[width=\linewidth,keepaspectratio]{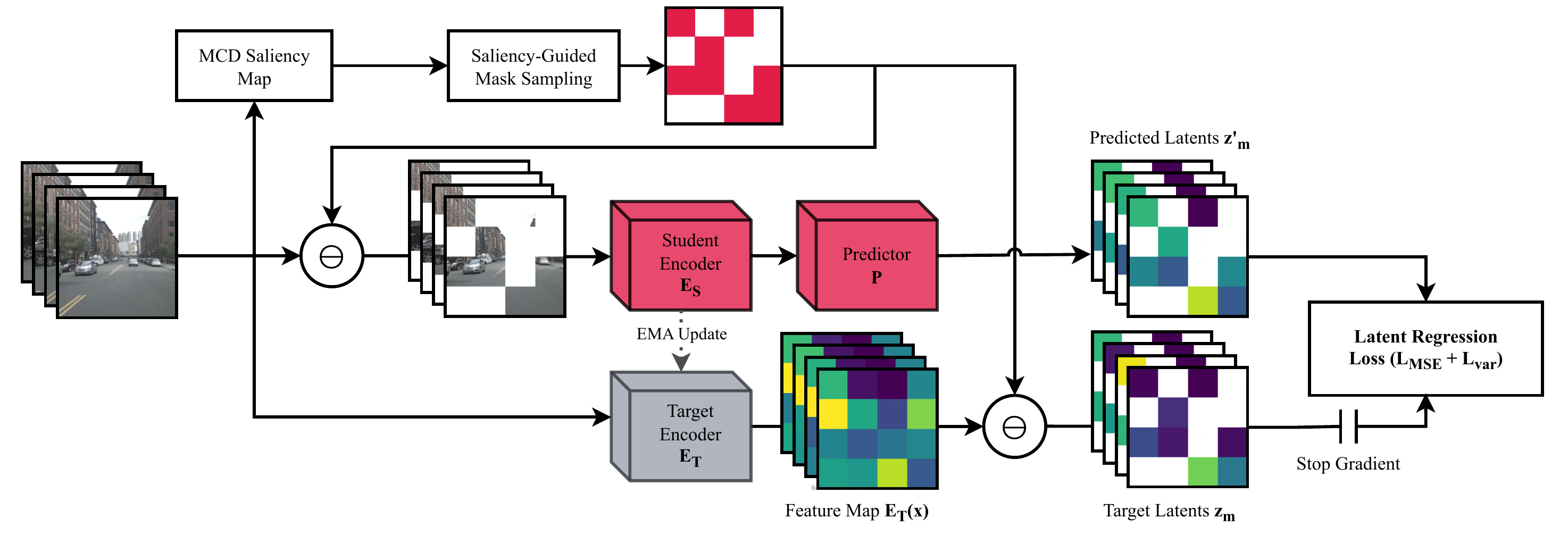}
    \caption{%
        Illustration of the \texttt{V-JEPA4A} training pipeline.
        Our saliency-guided mask generators first process a frame sequence to produce spatiotemporal masking tubes.
        The student encoder $E_S$ observes only the unmasked context tokens, while the target encoder $E_T$ processes the full (unmasked) clip to provide regression targets.
        $E_T$ can be operated in two modes: (\emph{i}) \textbf{EMA-updated}, where its weights are a slow exponential moving average of $E_S$ (following the original V-JEPA~\cite{vjepa}), or (\emph{ii}) \textbf{frozen}, where $E_T$ is initialized from a pre-trained checkpoint and kept fixed throughout training (following SALT~\cite{salt}).
        The lightweight predictor $P$ maps student context embeddings to predictions of the masked target embeddings, and the combined $\mathcal{L}_{\mathrm{MSE}} + \lambda\,\mathcal{L}_{\mathrm{var}}$ objective drives learning.
    }
    \label{fig:training_pipeline}
    \vspace{-0.3cm}
\end{figure*}

We build on the masked latent-prediction framework of V-JEPA~\cite{vjepa}, a ~\acrshort{ssl} approach that masks randomly sampled multi-block target regions and predicts their latent embeddings from the remaining visible context.
However, we propose to construct saliency-guided spatiotemporal maps that guide the random masking and tune them to preserve relevant regions in the visible context and mask less informative regions more aggressively, as can be seen in~\figref{fig:training_pipeline}

\subsection{Inherited Framework: Masked Latent Prediction}

We retain the masked latent-prediction architecture of V-JEPA~\cite{vjepa}, comprising a target encoder $E_T$, a student encoder $E_S$, and a lightweight predictor $P$.
A clip with $T$ frames at resolution $H{\times}W$ is tokenized into non-overlapping patches of size $t{\times}p{\times}p$ (by default we use $1{\times}14{\times}14$ patches). 
A masking policy $\mathcal{M}$ splits these tokens into visible context $\mathbf{x}_c$ and masked targets $\mathbf{x}_m$.
The student encoder processes only the visible context ($\mathbf{z}_c = E_S(\mathbf{x}_c)$) and the predictor estimates the latent features at the masked positions given only the unmasked tokens as context ($\hat{\mathbf{z}}_m = P(\mathbf{z}_c, \pi_m)$), where $\pi_m$ denotes positional queries for the masked tokens, constructed from the corresponding spatial and temporal positional embeddings.
The target encoder processes the full clip to generate regression targets $\mathbf{z}_m = \mathrm{LN}(E_T(\mathbf{x})[\mathbf{x}_m])$ of the masked target patches, where $LN$ denotes the layer norm operation.
We compare two target-encoder operating modes:
\begin{itemize*}
    \item \acrshort{ema} mode: $E_T$ follows V-JEPA~\cite{vjepa} and is updated as an exponential moving average of the student, providing slowly evolving self-supervised targets. 
    \item Frozen-teacher mode: $E_T$ is initialized from a pretrained checkpoint, e.g. DINOv3~\cite{dinov3}.
\end{itemize*} 

The training objective minimizes a regression loss augmented with a variance-collapse penalty:
\begin{align}
    \mathcal{L}
    =
    \underbrace{\|\hat{\mathbf{z}}_m - \mathbf{z}_m\|_2^2}_{\mathcal{L}_{\text{MSE}}}
    +
    \lambda\,
    \underbrace{\mathrm{ReLU}\!\left(1-\mathrm{std}(\hat{\mathbf{z}}_m)\right)}_{\mathcal{L}_{\text{var}}},
\end{align}
where the variance term keeps the predicted masked-token distribution from collapsing to a constant representation. 
We use $\lambda=0.1$ in all runs. 
Predicting masked tokens in latent space rather than reconstructing pixels, as in VideoMAE~\cite{videomae}, is well-suited for driving tasks, as the encoder can ignore certain photometric artifacts such as shadows and weather that are less informative for scene semantics, require no pixel decoder, and are aligned with the feature representations consumed by downstream perception heads.

\subsection{Saliency-Guided Masking}

We replace randomly sampled multi-block masking with a domain-specific saliency-guided policy. 
The key idea is that foreground regions, as specified by the saliency maps, should remain visible more often, while less informative background regions can be masked more aggressively. 
We evaluate this approach for the domain of ego-centric driving scenarios.

For the first pair of frames in each clip, we compute a per-pixel saliency map $\mathbf{S} \in [0,1]^{H \times W}$, aggregate it to patch resolution by average pooling, and normalize the magnitudes. 
Patches with saliency above threshold $\tau = 0.2 \cdot \max(\mathbf{S})$ are labeled foreground ($\mathcal{F}$), all others are labeled background. 
The relative thresholding is robust to clip-level intensity variations. 
We tune the threshold value using a small validation set and fix it for all runs. 
Threshold sensitivity remains a natural future ablation beyond the masking-ratio study in \tabref{tab:ablations}.
For patch $i$, the binary mask variable is sampled as
\begin{align}
    m_i \sim
    \begin{cases}
        \mathrm{Bern}(p_{\text{fg}}), & i \in \mathcal{F},\\
        \mathrm{Bern}(p_{\text{bg}}), & i \notin \mathcal{F},
    \end{cases}
\end{align}
where $m_i{=}1$ indicates that the patch is a masked target. 

If $\alpha=|\mathcal{F}|/N$ is the foreground-token fraction, then the expected overall target masking ratio is
\begin{align}
    r = \alpha\,p_{\text{fg}} + (1-\alpha)\,p_{\text{bg}}.
\end{align}
In the default run, we set $r=0.6$ and $p_{\text{fg}}=0.3$, then choose $p_{\text{bg}}=(r-\alpha p_{\text{fg}})/(1-\alpha)$ per clip. 
We employ this strategy to match the global masking ratio used by the random masking baselines.
The resulting token mask is reused for the full clip, with $T_{\text{clip}}{=}8$ frames in the default configuration. 

\begin{figure}[t]
    \centering

    \begin{subfigure}[t]{0.48\linewidth}
        \centering
        \includegraphics[width=\linewidth]{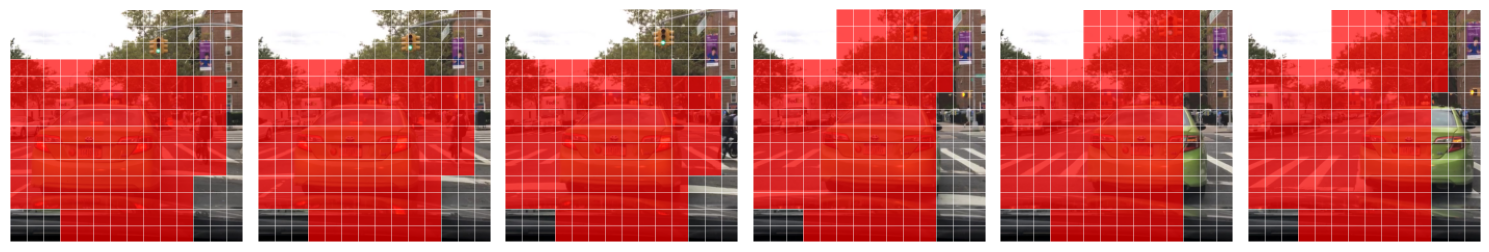}
        \caption{Random}
        \label{fig:mask_cmp_random}
    \end{subfigure}\hfill
    \begin{subfigure}[t]{0.48\linewidth}
        \centering
        \includegraphics[width=\linewidth]{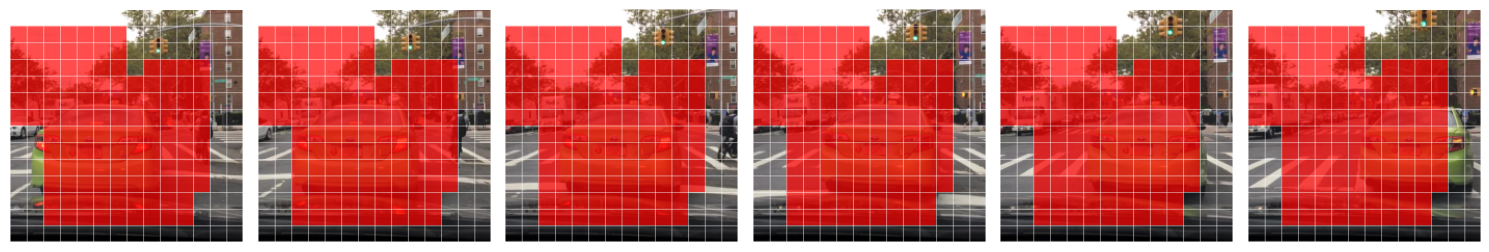}
        \caption{Sobel}
        \label{fig:mask_cmp_sobel}
    \end{subfigure}

    \vspace{0.3em}

    \begin{subfigure}[t]{0.48\linewidth}
        \centering
        \includegraphics[width=\linewidth]{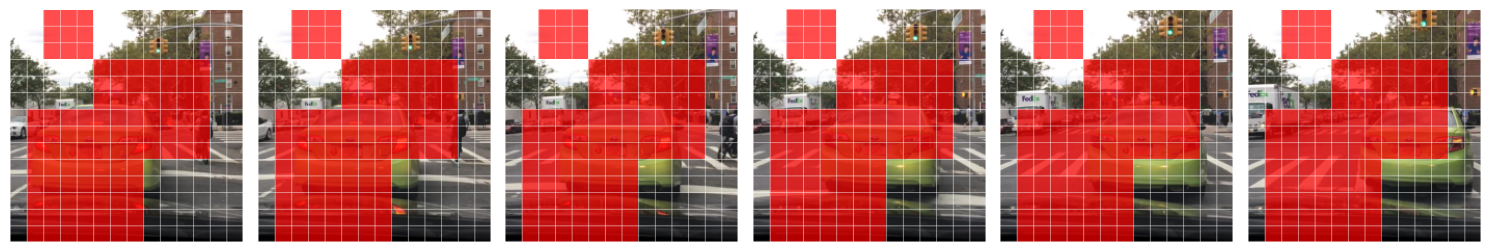}
        \caption{Optical Flow}
        \label{fig:mask_cmp_flow}
    \end{subfigure}\hfill
    \begin{subfigure}[t]{0.48\linewidth}
        \centering
        \includegraphics[width=\linewidth]{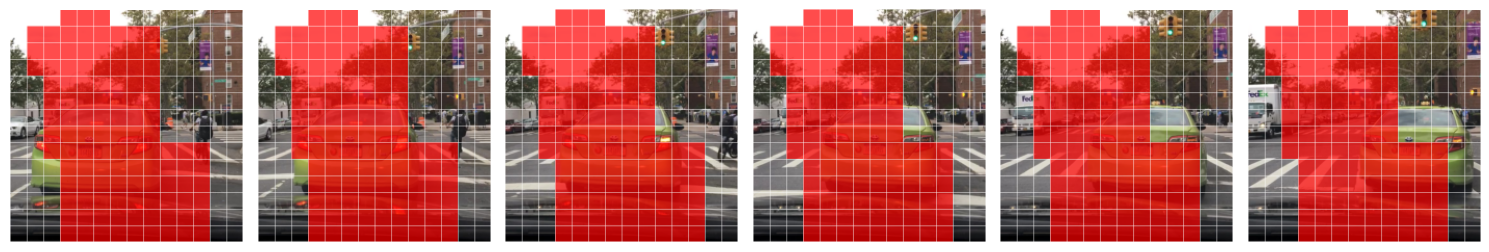}
        \caption{Motion-Compensated Differencing}
        \label{fig:mask_cmp_mcd}
    \end{subfigure}
    \vspace{-0.5em}
    \caption{Mask overlays for different masking strategies:
    \textbf{(a)}~Random,
    \textbf{(b)}~Sobel,
    \textbf{(c)}~Optical Flow,
    \textbf{(d)}~Motion-Compensated Differencing (\acrshort{mcd}).
    Red regions are held out for the student encoder during pre-training.
    }
    \label{fig:masks_comparison}
    \vspace{-0.3cm}
\end{figure}

We evaluate four saliency signals (depicted in \figref{fig:masks_comparison}): 
\begin{enumerate*}
    \item Sobel gradients, which respond to high-frequency edge structure but cannot distinguish static texture from moving objects. 
    \item Optical-flow magnitude, which captures motion but is contaminated by ego-motion in forward-facing driving video. 
    \item \acrfull{mcd}, which compensates camera ego-motion before differencing. 
    \item Combined Sobel\,+\,optical flow, which merges edge and flow, merging the strengths and weaknesses of both.
\end{enumerate*} 
We designed \acrshort{mcd} to suppress ego-motion induced background differences, thereby providing a clean foreground saliency signal in driving scenes.
We use it as the default masking strategy in \texttt{V-JEPA4A}.

\subsection{Motion-Compensated Frame Differencing}
\label{subsec:mcfd}
We define the \acrshort{mcd} saliency signal to isolate object motion from ego-motion.
Given two consecutive grayscale frames $\mathbf{G}_{t-1}, \mathbf{G}_t \in \mathbb{R}^{H \times W}$, we perform the following step-wise computation:
\begin{enumerate*}
    \item Computing dense optical flow using the Lucas-Kanade formulation over a local window.
    \item Using the estimated flow as a sampling grid, $\mathbf{G}_{t-1}$ is warped into the coordinate frame of $\mathbf{G}_t$ via bilinear interpolation, producing the compensated frame $\hat{\mathbf{G}}_{t-1}$.
    \item Computing the absolute residual $\mathbf{D} = \left|\mathbf{G}_t - \hat{\mathbf{G}}_{t-1}\right|$ to isolate motion that cannot be explained by ego-motion alone.
    \item Thresholding $\mathbf{D}$ with an adaptive threshold $\tau_{\text{adapt}} = 0.2 \cdot \max(\mathbf{D})$ to suppress weak responses from, e.g., parallax and sensor noise.
    \item The mask is pooled to patch resolution and normalized to $[0,1]$ to produce the saliency map.
\end{enumerate*}

\subsection{Relation to Prior Masking Schemes}
\label{subsec:relation_informed_masking}

V-JEPA random multi-block masking policy samples spatial blocks uniformly and treats all scene elements as interchangeable prediction targets.
\texttt{V-JEPA4A} preserves the latent-prediction objective but re-weights the probabilities for blocks to be samples in either foreground $p_{\text{fg}}$ and background regions $p_{\text{bg}}$ of the frame.
AdaMAE~\cite{adamae} learns token selection with an auxiliary network, which we found to be challenging to fine-tune in the latent prediction setup in V-JEPA, whereas our saliency policy is parameter-free.
MGMAE~\cite{mgmae} propagates masks with optical flow. However, in forward-facing driving videos, we found that ego-motion and object-scale changes can make flow-propagated masks drift and increase the effective masking ratio.
\acrshort{mcd} compensates for ego-motion before saliency estimation and reuses a fixed clip mask, thereby keeping the global masking ratio consistent across the clip.

\section{Experimental Evaluations}
\label{sec:experiments}
We evaluate \texttt{V-JEPA4A} on multiple downstream tasks relevant to autonomous driving.
Our code and trained checkpoints will be released upon acceptance.  

\subsection{Implementation Details}
\label{subsec:implementation_details}
We pre-train our models on publicly available driving datasets: nuScenes~\cite{nuscenes}, BDD100k~\cite{bdd100k}, KITTI~\cite{kitti}, and ImpromptuVLA~\cite{impromptuvla}, withholding the evaluation splits defined by dataset-specific benchmarks. The per-sample saliency masks are computed in parallel during data loading, so each run takes $\sim20$ hours on a node with 8$\times$ NVIDIA H200 GPUs. 
We evaluate the quality of the learned representations along four complementary axes by integrating the student encoder as a frozen backbone into task-specific perception stacks.
\begin{enumerate}[topsep=0pt]
    \item \textbf{Geometric awareness} is measured via monocular depth estimation on the KITTI-2015 benchmark~\cite{kitti}, reported as RMSE (lower is better). 
    We add a lightweight DPT~\cite{dpt} decoder head on top of the pre-trained encoder, and train only the decoder parameters for 20k iterations with an $\ell_1$ inverse-depth objective. 
    
    \item \textbf{Semantic richness} is assessed through semantic segmentation on Cityscapes~\cite{cityscapes}, reported as mean \acrshort{iou} (mIoU, higher is better). 
    Analogously to the depth estimation setup, we train a DPT~\cite{dpt} decoder head for 40 epochs with a cross-entropy objective function. 
    
    \item \textbf{Temporal consistency} is probed via appearance-similarity-based multi-object tracking on BDD100k~\cite{bdd100k}. 
    We extract crops from ground-truth detections and encode them as re-identification features, which are associated across frames with ByteTrack~\cite{bytetrack}.
    We report ID switches ($IDS$), which quantify the association errors across frames (lower values indicate better performance). 
    We omit compound \acrshort{mot} metrics such as MOTA and HOTA because the detections are shared across all models.
    
    \item \textbf{Semantic alignment} is evaluated using visual question answering on three driving-specific VQA benchmarks: NuScenes-MQA, OmniDrive, and ImpromptuVLA.
    For NuScenes-MQA~\cite{nuscenes-mqa}, we report yes/no accuracy, correct count accuracy for a prompted object category, a combined category/counting L1 error, and the relative L1 distance of referred objects to the ego vehicle.
    On OmniDrive~\cite{omnidrive}, we probe the model's ability to predict whether a simulated trajectory would be safe or violate traffic rules, reported by counterfactual reasoning accuracies.
    Finally, we evaluate open-loop trajectory prediction error for the ImpromptuVLA benchmark~\cite{impromptuvla} for a prediction horizon of up to 4 seconds.
    
    We integrate the frozen \texttt{V-JEPA4A} pre-trained ViT encoder into Qwen3VL-4B~\cite{qwen3vl} by replacing its original encoder and randomly reinitializing the merger parameters.
    We then train only the vision-language merger for 2 epochs using a combination of image and video-based scene description samples from the ImpromptuVLA~\cite{impromptuvla} and OmniDrive~\cite{omnidrive} datasets.
    Finally, we unfreeze the language model weights using PEFT~\cite{lora} with a learning rate of 1e-5 and a batch size of 64 for 5 epochs on all training samples.
    
\end{enumerate}

Full dataset statistics and training details are provided in the supplementary material.

\subsection{Benchmark Results}
\label{subsec:sota}

\begin{table}[t]
    \scriptsize
    \centering
    \caption{Benchmark results for self-supervised pre-trained vision transformers. 
    For depth estimation and semantic segmentation, the pretrained models were integrated as frozen vision encoders in a \acrshort{dpt} model. 
    For multi-object tracking, the ground truth detection boxes were used to pool feature representations for similarity-based matching across frames. Therefore, we report $IDS$ to isolates association quality. 
    Semantic segmentation was evaluated on Cityscapes~\cite{cityscapes}, depth prediction on KITTI-2015~\cite{kitti}, and multi-object tracking on BDD100k~\cite{bdd100k}.
    All results were obtained on the respective validation sets.}
    \label{tab:benchmark_evaluations}
    \setlength{\tabcolsep}{2.5pt}
    \renewcommand{\arraystretch}{1.1}
    \begin{tabular}{@{}lcccccc@{}}
        \toprule
        \textbf{Method} & 
        \textbf{Model} & 
        \shortstack[c]{\textbf{Patch}\\\textbf{size}} & 
        \shortstack[c]{\textbf{Input}\\\textbf{resolution}} & 
        \shortstack[c]{\textbf{Sem. seg.}\\\textbf{mIoU [\%] $\uparrow$}} & 
        \shortstack[c]{\textbf{Depth}\\\textbf{RMSE [m] $\downarrow$}} & 
        \shortstack[c]{\textbf{MOT}\\\textbf{[IDS] $\downarrow$}} \\ \midrule
        DINOv2~\cite{dinov2}      & ViT-B          & 1x14x14             & 224x224             & 66.5                                  & 5.31                                      & 10550                               \\
        DINOv2~\cite{dinov2}      & ViT-L          & 1x14x14             & 224x224             & 69.3                                  & \textbf{3.57}                                      & 10283                               \\
        DINOv3~\cite{dinov3}      & ViT-L          & 1x16x16             & 512x512             & \underline{72.0}                                  & 4.10                                      & 9877                                \\
        OpenCLIP~\cite{clip}      & ViT-B          & 1x14x14             & 224x224             & 54.7                                  & 5.15                                      & 14098                               \\
        V-JEPA~\cite{vjepa}       & ViT-L          & 2x16x16             & 224x224             & 68.2                                  & 4.58                                      & 12002                               \\
        VideoMAE~\cite{videomae}  & ViT-B          & 1x14x14             & 518x518             & 42.1                                  & 8.01                                      & 14208                               \\ \midrule
        \shortstack[l]{V-JEPA4A (random)} & ViT-B          & 1x14x14             & 518x518             & 48.4                                  & 6.90                                      & 10047                               \\
        V-JEPA4A                  & ViT-B          & 1x14x14             & 518x518             & 64.6                                  & 4.52  	& \underline{9134}                                \\
        \shortstack[l]{V-JEPA4A (12 frames)} & ViT-B          & 1x14x14             & 518x518            & 69.3                                  & 3.97                                      & 9205                                \\ \midrule
        \shortstack[l]{V-JEPA4A (random, DINOv3)} & ViT-L          & 1x16x16             & 512x512             & 51.9                                  & 4.24                                      & 9906                                \\

        \shortstack[l]{V-JEPA4A (DINOv3 teacher)} & ViT-L          & 1x16x16             & 512x512             & \textbf{73.2}                         & \underline{3.75}                                      & \textbf{9018}                       \\ \bottomrule
    \end{tabular}
    \vspace{-0.6cm}
\end{table}

\tabref{tab:benchmark_evaluations} compares \texttt{V-JEPA4A} against a set of established \acrshort{ssl} methods, including DINOv2~\cite{dinov2}, DINOv3~\cite{dinov3}, OpenCLIP~\cite{clip}, VideoMAE~\cite{videomae}, and the original V-JEPA~\cite{vjepa} using random multi-block masking. 
\figref{fig:spider_plot} provides a compact visual summary of these comparisons.

{\parskip=2pt
\noindent\textit{Multi-Object Tracking}: Our most considerable improvement over the baselines occurs in multi-object tracking, 
where \texttt{V-JEPA4A} (ViT-B, $518\times518$) reduces the number of ID switches by $24\%$ from 12002 (V-JEPA) to 9134.
Scaling \texttt{V-JEPA4A} with a DINOv3 distillation teacher and a ViT-L backbone improves tracking further to 9018 IDS, achieving the best result among all compared methods. 
We attribute these performance improvements to the dedicated handling of foreground objects that ensures these are visible throughout the training sequence, allowing the encoder to learn temporally consistent representations.}

{\parskip=2pt
\noindent\textit{Semantic Segmentation}: \texttt{V-JEPA4A} achieves 64.6 mIoU, which is below image-based models of comparable size (DINOv2 ViT-B achieving 66.5 mIoU). 
This can be attributed to the trade-off between temporal objectives and fine-grained spatial detail.
Increasing the input resolution to $518\times518$ and extending the training sequence to 12 frames substantially closes this gap, reaching 69.3 mIoU and performing on par with DINOv2 ViT-L (69.3 mIoU) while using a smaller backbone.
The DINOv3-teacher variant achieves 73.2 mIoU, surpassing all compared methods in \tabref{tab:benchmark_evaluations}.
These results indicate that with sufficient resolution and a strong spatial teacher, \texttt{V-JEPA4A} is competitive with state-of-the-art approaches for dense semantic prediction, even when trained on orders of magnitude less data, drawn only from the driving domain.}

{\parskip=2pt
\noindent\textit{Depth Estimation}: Our \texttt{V-JEPA4A} pre-trained on clips of 8 frames achieves an RMSE of \SI{4.52}{\meter}, whereas the 12-frame variant reduces the error further to \SI{3.97}{\meter}, performing on par with DINOv3 using a larger ViT-L (4.10) architecture.
DINOv2 ViT-L notably achieves the best overall depth score (\SI{3.57}{\meter}), suggesting that contrastive pre-training on massive data also provides strong geometric cues.}

\begin{figure*}[t]
    \centering

    \begin{minipage}[t]{0.98\linewidth}
        \centering
        \includegraphics[width=\linewidth]{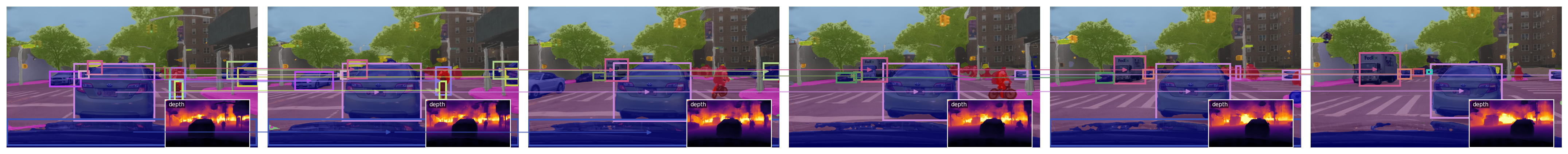}
    \end{minipage}

    \vspace{0.5em}

    \begin{minipage}[t]{0.98\linewidth}
        \centering
        \includegraphics[width=\linewidth]{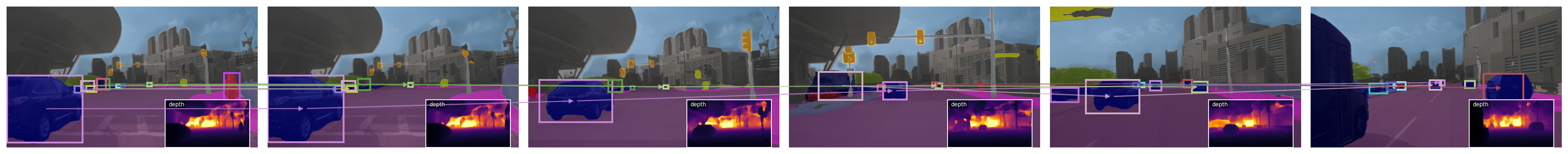}
    \end{minipage}

    \caption{Representative downstream-task visualizations over six-frame sequences.
    Each composite overlays semantic segmentation, monocular depth estimation, and multi-object tracking predictions on BDD100K scene b1c9c847-3bda4659 (top) and a nuScenes scene 0321 (bottom).}
    \label{fig:downstream_demo_main}
    \vspace{-0.4cm}
\end{figure*}
\begin{table*}[t]
    \scriptsize
    \centering
    \caption{
    Comparison of model performances across three driving-related visual question answer benchmarks. 
    \textbf{Bold} = best, \underline{underline} = second best within each column. 
    The top section reports benchmark-specific reference results taken from the cited works.
    The controlled comparison is between the Qwen3-based models in the lower section, which only differ in the \texttt{V-JEPA4A} pre-training for the vision encoder.    
    }
    \label{tab:vlm_benchmarks}
    \setlength{\tabcolsep}{2.7pt}
    \renewcommand{\arraystretch}{1.05}
    \begin{tabular}{@{}lccccccccccc@{}} \toprule
        \textbf{Method} & \multicolumn{3}{c}{\textbf{NuScenes MQA}} & \multicolumn{4}{c}{\textbf{OmniDrive}} & \multicolumn{4}{c}{\textbf{ImpromptuVLA}} \\
        \cmidrule(lr){2-4} \cmidrule(lr){5-8} \cmidrule(lr){9-12}
        & \shortstack[c]{Y/N\\Acc.} & \shortstack[c]{Cat.\\Count} & \shortstack[c]{Rel. dist.\\L1} & Safe & \shortstack[c]{Red\\light} & Coll. & \shortstack[c]{Driv.\\area} & 1s & 2s & 3s & 4s \\ \midrule
        OPT-6.7B~\cite{nuscenes-mqa} & \textbf{0.82} & \underline{0.32} & 4.21 & -- & -- & -- & -- & -- & -- & -- & -- \\
        Omni-L (7B)~\cite{omnidrive} & -- & -- & -- & \underline{72.1} & \underline{59.2} & \textbf{34.3} & 49.1 & -- & -- & -- & -- \\
        Impromptu-VLA (7B)~\cite{impromptuvla} & -- & -- & -- & -- & -- & -- & -- & \textbf{0.10} & \underline{0.31} & \underline{0.65} & \underline{1.11} \\ \midrule
        \shortstack[l]{Qwen3-VJEPA (random)} & 0.71 & 0.29 & \underline{4.07} & 64.7 & 31.7 & 25.2 & \textbf{58.6} & 0.11 & 0.38 & 0.73 & 1.30 \\
        \shortstack[l]{Qwen3-VJEPA4A (MCD)} & \underline{0.78} & \textbf{0.40} & \textbf{2.80} & \textbf{80.3} & \textbf{62.4} & \underline{29.8} & \underline{54.0} & 0.11 & \textbf{0.29} & \textbf{0.58} & \textbf{1.02} \\ \bottomrule
    \end{tabular}
    \vspace{-0.3cm}
\end{table*}

\figref{fig:downstream_demo_main} presents qualitative overlays of the downstream predictions from two evaluation sequences on the BDD100K and the nuScenes datasets.
Across both scenes, the learned representation supports coherent semantic segmentation, stable depth maps, and temporally consistent tracking trajectories over the entire clip.

{\parskip=2pt
\noindent\textit{Visual question answering}: 
The VQA results are shown in \tabref{tab:vlm_benchmarks}, where the upper block reports benchmark-specific reference models that serve as contextual baselines because they differ in scale, training data, and answer formatting. 
The controlled comparison is between \texttt{Qwen3-VJEPA} with random masking and \texttt{Qwen3-VJEPA4A}, which differ only in the vision encoder but use the same training recipe and unified answer-tag format.
In this setting, \texttt{Qwen3-VJEPA4A} obtains the best mean performance, improving OmniDrive mean accuracy across all question types from 45.1 to 56.6, and reducing the Impromptu-VLA 4s trajectory L2 error from 0.63 to 0.50 for the 4-second planning horizon.}

\subsection{Ablation Study}  
\label{subsec:ablation}  

\begin{table}[t]
    \scriptsize
    \centering
    \caption{Single-factor ablations. The baseline (Run~1, shaded) uses ViT-B, $518{\times}518$ , 8 frames, $1{\times}14{\times}14$ patches, 
    MCD masking at ratio 0.6, temporal PE, and an EMA-updated target encoder. 
    Each group varies one factor.}
    \label{tab:ablations}
    \setlength{\tabcolsep}{3.5pt}
    \begin{tabular}{@{}lcl rrr@{}} \toprule
       & & & \textbf{BDD100k} & \textbf{Cityscapes} & \textbf{KITTI} \\
        \cmidrule(lr){3-3} \cmidrule(lr){4-4} \cmidrule(lr){5-5}
        & Run & Configuration & IDS $\downarrow$ & mIoU $\uparrow$ & RMSE $\downarrow$ \\ \midrule
        \rowcolor{black!8}\multirow{5}{2.5cm}{Masking strategy (ratio\,=\,0.6)} & 1 & MCD (baseline)    & 9134  & 64.6 &  4.52 \\
        & 2                    & Random            & 10047  & 48.4 & 6.90 \\
        & 3                    & Sobel             & 10358 & 48.2 & 6.14 \\
        & 4                    & Flow              & 9712  & 53.7 & 4.24 \\
        & 5                    & Sobel+Flow        & 9902  & 52.4 & 5.10 \\ \midrule
        \multirow{3}{*}{Masking ratio (MCD)} & 6                    & 0.2               & 10170 & 69.7 & 6.91 \\
        & 7                    & 0.5               & 9882  & 73.1 & 5.70 \\
        & 8                    & 0.8               & 10024 & 62.8 & 6.38 \\ \midrule
       \multirow{2}{2.5cm}{Foreground masking probability $p_{fg}$} & 9                    & 0.1               & 9601  & 61.0 & 5.28 \\
        & 10                   & 0.5               & 9720  & 60.3 & 5.24 \\ \midrule
        \multirow{2}{*}{Temporal context} & 11                   & 4 frames                      & 9507  & 60.3            & 4.74  \\
        & 12                   & 12 frames                     & 9205  & \textbf{69.3}   & \underline{\textbf{3.97}} \\ \midrule
        Temporal PE & 13                   & no temporal PE                & \underline{9122}    & 51.4            & 4.82  \\ \midrule
        \multirow{2}{*}{Input resolution} & 14                   & res.\ $224{\times}224$   & 9086  & 52.6          & 7.02 \\
        & 15                   & res.\ $384{\times}384$   & 10109 & 57.4          & 5.19 \\ \midrule
        Model size & 16                   & ViT-L                & 9204  & 58.9          & \textbf{3.95}3.95 \\ 
\bottomrule
    \end{tabular}
    \vspace{-0.4cm}
\end{table} 

To quantify the influence of each design choice and gain insights for adapting \texttt{V-JEPA4A}, we summarize ablation results in \tabref{tab:ablations}. The baseline (Run~1) uses a ViT-B encoder, 8 frames, masking ratio 0.6, \acrshort{mcd} saliency estimation, resolution $518\times518$, and additive temporal positional encoding.\looseness=-1

{\parskip=2pt
\noindent\textit{Masking Strategy}: Among the methodological choices, the masking strategy has the largest single-factor effect.
Replacing \acrshort{mcd} with \emph{random} masking (Run~2) severely degrades segmentation to 48.4 mIoU, while optical-flow masking (Run~4) yields a partial improvement (53.7 mIoU), but it falls short of \acrshort{mcd} with 64.6 mIoU.
Naively combining Sobel edges and optical flow (Run~5) does not improve on either signal alone, indicating that the sum of saliency signals still suffers from their complementary failure modes.
In terms of masking ratio, the baseline ratio of 0.6 (Run~1) provides the best balance for tracking and depth. 
Lower ratios (Runs~6--7) improve mIoU but degrade IDS and RMSE, while aggressive masking at 0.8 (Run~8) degrades all metrics relative to the baseline.}

{\parskip=2pt
\noindent\textit{Foreground Masking Probability}: Reducing the foreground masking probability from 0.3 (Run~1) to 0.1 (Run~9) causes an increase in the depth RMSE (+0.76m) and tracking error (+5\% more ID switches), indicating that weak masking of foreground affects both dense and object-level feature representations.
Increasing the masking probability for a foreground patch to 0.5 (Run~10) still underperforms the baseline, primarily in depth estimation accuracy, 
indicating that moderate foreground masking is important for balancing geometrical and semantic feature representations.}

{\parskip=2pt
\noindent\textit{Temporal Context}: Increasing the clip from 8 to 12 frames (Run~12) yields the best segmentation and depth while slightly degrading multi-object tracking performance. 
A clip length of 4 frames per training sample (Run~11) degrades all three performance metrics, suggesting that longer temporal context enables learning more expressive feature representations. 
However, increasing the clip length from 8 to 12 frames adds substantial computational requirements. 
Having no temporal positional encoding during training (Run~13) reduces both segmentation and depth accuracy, but slightly improves MOT performance, likely because it produces position-invariant features that better match the re-identification features used by the tracker.
}

{\parskip=2pt
\noindent\textit{Architecture}: The image resolution is a key contributing factor, lowering the image resolution from $518{\times}518$ to $224{\times}224$ (Run~14) reduces performance on dense prediction tasks, the segmentation mIoU drops sharply from 64.6 to 52.6 mIoU.
In contrast, multi-object tracking appears less sensitive to the input resolution, presumably because it mainly relies on object-level feature representations.
Scaling to ViT-L (Run~16) degrades segmentation and tracking while slightly improving depth, suggesting that larger models may require longer training schedules.
}

\subsection{Discussion of Limitations}
\label{subsec:limitations}

Our model is a specialized encoder for the driving domain. In our experiments, we intentionally trade the broad coverage of internet-scale general-purpose backbones such as DINOv3 for an inductive bias tailored to safety-critical motion and egocentric driving scenes. 
This makes the approach effective for driving-scene representation learning, but not a replacement for large foundation models with broader visual coverage.
Thus, our results validate domain-aware masking for driving, while its effectiveness under larger and more diverse pre-training remains open.
The VLM evaluations should therefore be interpreted as controlled comparisons of visual encoders, not as benchmarks against all specialized models.

\section{Conclusion}
\label{sec:conclusion}
We introduced \texttt{V-JEPA4A}, a driving-specific variant of V-JEPA that replaces content-agnostic random multi-block masking with \acrfull{mcd}-based saliency-guided masking. 
By aligning masked latent prediction with semantically and temporally informative regions, \texttt{V-JEPA4A} preserves safety-critical cues as context during pre-training and learns representations better suited to egocentric driving perception. 
Evaluations across four autonomous-driving tasks yield two main findings.
First, domain-aware masking produces representations competitive with models trained on orders of magnitude more data, despite using only publicly available driving videos. 
Second, in our setting, the saliency signal defining the pretext task has a stronger effect than model or data scale. 
Modifying only the masking policy yields larger gains than scaling the backbone from ViT-B to ViT-L. 
These results show that lightweight domain knowledge, injected through the self-supervised objective rather than labels, losses, or architectural changes, can substantially improve representation quality in safety-critical driving scenarios. 
At the same time, \texttt{V-JEPA4A} remains a domain-specialized encoder rather than a general-purpose visual foundation model. 
Its benefits are demonstrated in controlled driving-domain evaluations, while the generalization of \acrshort{mcd}-guided masking to larger, mixed-domain pre-training corpora, as well as its use for fine-tuning pre-trained visual foundation models, remains important future work.


\bibliographystyle{splncs04}
\bibliography{references}

\clearpage

\title{Mask What Matters: Saliency-Guided Video Self-Supervised Learning for Autonomous Driving \newline \textit{Supplementary Material}}

\author{Christopher Lang\inst{1,2} \and
Alexander Braun\inst{1} \and
Abhinav Valada\inst{2}}

\institute{Robert Bosch GmbH, Stuttgart, Germany \and
Robot Learning Lab, University of Freiburg, Freiburg, Germany\\
\email{\{lang,valada\}@cs.uni-freiburg.de}}

\titlerunning{Supplementary - Saliency-Guided Video SSL for Autonomous Driving}
\authorrunning{C. Lang et al.}

\maketitle

\appendix
\renewcommand{\theHsection}{app.\thesection}
\renewcommand{\theHsubsection}{app.\thesubsection}

\section{Implementation Details}
\label{app:implementation}

This section provides additional details on the datasets (\secref{app:datasets}), hyperparameter settings for pre-training (\secref{app:pretrain_task}) and downstream tasks (\secref{app:downstream_integration}), and the computational resources used in our experiments (\secref{app:compute_resources}).

\subsection{Dataset Statistics}
\label{app:datasets}

\tabref{tab:datasets} summarizes datasets used in our experiments, including splits, frame counts, frame rates, and resolutions.
\tabref{tab:datasets_pretrain} provides additional statistics for the pre-training datasets.
We build the clips used for training by subsampling the raw videos to a uniform frame rate of 2\,FPS and grouping them into 16-frame clips, resulting in clip lengths of 8 seconds.
We chose a frame rate of 2\,FPS as the greatest common factor among our data sources. The clips are non-overlapping.
\vspace{-2pt}  
\begin{table}[H]
    \footnotesize
    \centering  
    \caption{Overview of the datasets used in pre-training (\textbf{PT}) and downstream fine-tuning (\textbf{FT}). 
    FPS values correspond to the raw video frame rate; we subsample where necessary to obtain a uniform temporal resolution.
    The splits refer to labeled sequences, while the number of sequences counts all available video sequences, including those without labels.
    }  
    \label{tab:datasets}  
    \begin{tabular}{lccccc}  
        \toprule  
        \multirow{2}{*}{Dataset} & Split & \# Seqs  & FPS & Resolution & Usage\\  
        \cmidrule{2-6}  
         & train/val/test &  &   &  &  \\  
        \midrule  
        KITTI~\cite{kitti}         & 28 / 0 / 11 & 170 &  10 & $1242\times375$ & PT / FT \\  
        BDD100k~\cite{bdd100k}           & 70 / 10 / 20 & 1\,100 & 30 & $1280\times720$ & PT / FT \\  
        nuScenes~\cite{nuscenes}      & 700 / 150 / 150 & 1\,000 & 12 & $1600\times900$ & PT / FT \\
        ImpromptuVLA~\cite{impromptuvla}$^\dagger$ & 56 / 14 / 0 & 1.1 & 2 & Variable & PT / FT \\
        Cityscapes~\cite{cityscapes}    & 23 / 1 / 2  & 75  & 2.5 & $2048\times1024$ & FT \\    
        \bottomrule
        \multicolumn{6}{l}{\scriptsize $^\dagger$Excludes nuScenes and KITTI clips already listed above.}  
    \end{tabular}  
\end{table}

\begin{table}[ht]
    \centering
    \footnotesize
    \caption{Pretraining data Hours are clip-hours computed as clips $\times$ frames per clip / FPS.}
    \begin{tabular}{lrrrr}
        \toprule
        Dataset       & Clips  & Frames/clip & FPS & Hours  \\ \midrule
        BDD100K       & 58,610 & 16          & 2.0 & 130.24 \\
        nuScenes      & 25,042 & 16          & 2.0 & 55.65  \\
        KITTI         & 284    & 16          & 2.0 & 2.40   \\
        ImPromptU VLA$^\dagger$ & 15,715 & 16          & 2.0 & 34.92  \\
        \bottomrule
        \multicolumn{5}{l}{\scriptsize $^\dagger$Excludes nuScenes and KITTI clips already listed above.}  
    \end{tabular}
    \label{tab:datasets_pretrain}
\end{table}

\subsection{Pre-training Task}
\label{app:pretrain_task}

\tabref{tab:pretrain_hyperparameters} provides a detailed list of pre-training hyperparameters.
In \tabref{tab:masking_latency}, we report latency benchmarks for different mask computation strategies on an Intel i9-9980XE CPU running at 3.00 GHz. Since mask generation is performed within the dataloader workers, these measurements primarily reflect the additional computational overhead introduced by each method. The reported values correspond to the average mask computation time per training sample (an 8-frame clip at a resolution of $518{\times}518$), averaged over 200 samples.
 
\begin{table}[!htbp]
    \scriptsize
    \centering
    \caption{%
        V-JEPA4A encoder architecture and pre-training hyperparameters.%
    }
    \label{tab:pretrain_hyperparameters}
    \setlength{\tabcolsep}{5pt}
    \renewcommand{\arraystretch}{1.0}
    \begin{tabular}{@{}l l p{0.52\textwidth}@{}}
    \toprule
    \textbf{Group} & \textbf{Hyperparameter} & \textbf{Value} \\

    \midrule
    \multicolumn{3}{@{}l}{Shared ViT encoder architecture} \\
    \cmidrule(l){1-3}
    & Image size                & $518{\times}518$ \\
    & Patch size                & $14$ \\
    & Architecture              & ViT-B \\
    & Class token               & No \\
    & Positional embedding      & Multi-resolution RoPE (MRoPE) \\
    & Drop-path rate            & $0.1$ \\

    \midrule
    \multicolumn{3}{@{}l}{V-JEPA4A pre-training} \\

    \cmidrule(l){1-3}
    \multicolumn{3}{@{}l}{\quad\textit{Video sampling}} \\
    & Num.\ frames $T$            & $8$ \\
    & Tubelet size                & $1{\times}14{\times}14$ \\

    \cmidrule(l){2-3}
    \multicolumn{3}{@{}l}{\quad\textit{Saliency-guided masking (Sobel)}} \\
    & Foreground sampling $p_{\text{fg}}$ & $0.3$ \\
    & Masking ratio                       & $0.6$ \\
    & Saliency threshold $\tau$           & $0.2 \cdot \max(\mathbf{S})$ \\
    & Min block size $(h, w)$             & $(2, 2)$ patches \\
    & Num.\ mask blocks                   & $8$ \\
    & Spatial scales                      & Progressive: $[0.05,\!0.05]$ to $[0.2,\!0.3]$ \\
    & Temporal scales                     & $[1.0, 1.0]$ (all blocks) \\
    & Aspect ratios                       & $[0.5,\!1.5]$ / $[0.5,\!2.0]$ (alternating)\\

    \cmidrule(l){2-3}
    \multicolumn{3}{@{}l}{\quad\textit{Predictor}} \\
    & Predictor depth           & $6$ \\
    & Predictor embed dim       & $768$ (= encoder $d$) \\

    \cmidrule(l){2-3}
    \multicolumn{3}{@{}l}{\quad\textit{Loss}} \\
    & Loss exponent             & $1.0$ \\
    & Variance regularization $\gamma$ & $0.1$ \\

    \cmidrule(l){2-3}
    \multicolumn{3}{@{}l}{\quad\textit{Target encoder / EMA}} \\
    & EMA start epoch           & $0$ \\
    & Momentum schedule         & $[0.99925, 0.99925]$ \\
    & Teacher mode              & EMA-updated target \textbf{or} frozen DINOv3 ViT-L \\

    \cmidrule(l){2-3}
    \multicolumn{3}{@{}l}{\quad\textit{Optimisation (warmup--cosine--cooldown)}} \\
    & Optimizer                 & AdamW, $\beta=(0.9, 0.95)$, $\varepsilon=10^{-8}$ \\
    & Learning rate             & $10^{-5} \!\to\!5{\times}10^{-4} \!\to\! 10^{-6}$ \\
    & Batch size                & $128$ \\
    & Total steps               & $147{,}000$ \\
    & Warmup phase              & $12{,}000$ steps \\
    & Weight decay              & $0.04$ \\
    & Gradient clipping         & $1.0$ (max norm) \\

    \cmidrule(l){2-3}
    \multicolumn{3}{@{}l}{\quad\textit{Data augmentation}} \\
    & Random resize scale       & $[0.3, 1.0]$ \\
    & Aspect-ratio range        & $[0.75, 1.35]$ \\
    & Normalization             & ImageNet (mean / std) \\
    \bottomrule
    \end{tabular}
\end{table}

\begin{table}[H]
  \centering
  \footnotesize
  \caption{%
    Latency overhead of saliency-guided masking strategies relative to
    random masking (ViT-B/14, $518{\times}518$, $T{=}8$ frames).
  }
  \label{tab:masking_latency}
  \setlength{\tabcolsep}{5pt}
    \begin{tabular}{lr} \toprule
                                    & \textbf{Mask comp. [ms]} \\ \midrule
    \textsc{Random}                 & 31.4                     \\
    \textsc{Sobel}                  & 59.7                     \\
    \textsc{Optical Flow}           & 78.8                     \\
    \textsc{Mot.-Comp. Diff.}       & 88.4                     \\
    \textsc{Sobel $+$ Flow}         & 93.5                     \\ \bottomrule
    \end{tabular}
\end{table}

\subsection{Downstream Task Integration}
\label{app:downstream_integration}

For downstream evaluation, we freeze the parameters of the pre-trained encoder.
During task-specific fine-tuning, we add a task-specific head and update only its parameters.
We describe the head implementations below.
The default hyperparameter settings are reported in~\tabref{tab:downstream_hyperparameters}.

\begin{table}[!htbp]
    \scriptsize
    \centering
    \caption{%
        Downstream task hyperparameters.  All tasks use the shared frozen
        V-JEPA4A encoder from \cref{tab:pretrain_hyperparameters};
        only task-specific heads and adapters are trained.%
    }
    \label{tab:downstream_hyperparameters}
    \setlength{\tabcolsep}{5pt}
    \renewcommand{\arraystretch}{1.0}
    \begin{tabular}{@{}l l p{0.52\textwidth}@{}}
    \toprule
    \textbf{Task} & \textbf{Hyperparameter} & \textbf{Value} \\

    \midrule
    \multicolumn{3}{@{}l}{(A) Cityscapes semantic segmentation (DPT)} \\
    \cmidrule(l){1-3}
    & Num.\ classes             & $19$ \quad(\texttt{ignore\_index}$=255$) \\
    & Loss                      & Cross-entropy \\
    & Num.\ frames / tubelet    & $1$ / $1$ \\
    & Batch size                & $32$ \\
    & Learning rate $\eta$      & $10^{-6} \!\to\!2{\times}10^{-4} \!\to\! 10^{-5}$ \\
    & Weight decay              & $0.04$ \\
    & Gradient clipping         & $1.0$ \\
    & Warmup / total epochs     & $1$ / $10$ \\
    & Random resize scale       & $[0.5, 2.0]$ \\
    & Aspect-ratio range        & $[0.75, 1.35]$ \\

    \midrule
    \multicolumn{3}{@{}l}{(B) KITTI monocular depth estimation (DPT)} \\
    \cmidrule(l){1-3}
    & Decoder head              & DPT \\
    & Output dim                & $1$ (continuous depth) \\
    & Loss                      & Sparse RMSE \\
    & Max depth                 & $80.0$\,m \\
    & Num.\ frames / tubelet    & $1$ / $1$ \\
    & Optimisation              & Same as (A) \\

    \midrule
    \multicolumn{3}{@{}l}{(C) Evaluation: multi-object tracking (zero-shot)} \\
    \cmidrule(l){1-3}
    & Tracking mode             & ReID features \\
    & Init.\ track threshold    & $0.7$ \\
    & Detection score thresholds & High: $0.6$, Low: $0.1$ \\
    & Num.\ tentative frames    & $3$ \\
    & Max frames to retain      & $30$ \\
    
    \midrule
    \multicolumn{3}{@{}l}{(D) Fine-tuning: Video-Language Model (VLM)} \\

    \cmidrule(l){1-3}
    \multicolumn{3}{@{}l}{\quad\textit{Architecture}} \\
    & Base LLM                  & Qwen3-VL-4B-Instruct \\
    & Vision encoder            & V-JEPA4A ViT (frozen) \\
    & Merger reinitialisation   & Xavier \\
    & LoRA rank $r$ / $\alpha$  & $8$ / $32$ \\
    & LoRA dropout              & $0.1$ \\
    & LoRA targets              & \texttt{q\_proj}, \texttt{k\_proj}, \texttt{v\_proj}, \texttt{o\_proj} \\

    \cmidrule(l){2-3}
    \multicolumn{3}{@{}l}{\quad\textit{Training}} \\
    & Max text length           & $2048$ tokens \\
    & Per-device batch          & $16$ \\
    & Gradient accumulation     & $4$ \quad(eff.\ batch = 64) \\
    & Warmup steps              & $100$ \\
    & LR scheduler              & Cosine \\
    & Weight decay              & $0.01$ \\
    & Max gradient norm         & $1.0$ \\
    & Precision                 & \texttt{bf16} \\

    \bottomrule
    \end{tabular}
\end{table}

\paragraph{Dense prediction (DPT).}
For semantic segmentation on Cityscapes and monocular depth estimation on KITTI-2015, we adopt the Dense Prediction Transformer (DPT)~\cite{dpt} architecture.
Multi-scale features are extracted from vision encoder layers $\{0, 2, 5, 11\}$ and reassembled into a four-level feature pyramid with reassembly factors $[4, 2, 1, 0.5]$ and neck channel widths $[96, 192, 384, 768]$.
The DPT patch-embedding layer is warm-started from the parameters of the pre-trained student encoder.
The ViT backbone remains frozen; only the DPT embedding layer, reassembly neck, and task-specific prediction head are trained.

\paragraph{Multi-object tracking (ByteTrack).}
We evaluate the frozen encoder in a zero-shot setting on the BDD100K multi-object tracking benchmark. 
Given ground-truth detection bounding boxes, each frame is passed once through the pre-trained student encoder. 
Patch tokens whose spatial indices fall inside a detection box are mean-pooled to obtain a one-dimensional re-identification feature vector for each detection. 
The tracker implementation follows the ByteTrack~\cite{bytetrack} protocol.

\paragraph{Video question answering (Qwen3-VL + LoRA).}
We replace the native Qwen3-VL vision tower with the frozen \texttt{V-JEPA4A} encoder and randomly initialize the patch-merger and deep-stack merger layers. 
Each merger aggregates groups of $2{\times}2$ patch tokens by concatenating them along the feature dimension ($4d$ dimensions) and passing them through a two-layer MLP that maps them into the text-token embedding space.
While the patch merger operates on the final output patch tokens, the deep-stack mergers produce multi-scale visual features from intermediate patch tokens, which are concatenated with the patch-merger output.

The ViT backbone is fully frozen; the trainable components are
\begin{enumerate*}[label=(\alph*)]
    \item the patch- and deep-stack merger MLPs, which receive a $10{\times}$ learning-rate multiplier, and
    \item rank-$8$ LoRA adapters on the language model's projection layers
\end{enumerate*}
(see~\cref{tab:downstream_hyperparameters}, group~D).

\subsection{Compute Resources}
\label{app:compute_resources}

All self-supervised pre-training runs are conducted on 8$\times$ NVIDIA H200 GPUs, with wall-clock training times ranging from approximately 20 to 26 hours depending on the model configuration (\emph{e.g.}, backbone size, input resolution, and clip length).

DPT-based downstream evaluations for depth estimation and semantic segmentation take roughly 4 hours on 4$\times$ H200 GPUs per run. 
The VLM fine-tuning pipeline, including two-stage merger pre-training and LoRA instruction tuning, takes approximately 8 hours on 4$\times$ H200 GPUs per run.

\section{Mask Generator Visualizations}
\label{app:mask_generators}

\begin{figure*}[t]
    \centering

    \begin{subfigure}[t]{\linewidth}
        \centering
        \includegraphics[width=\linewidth]{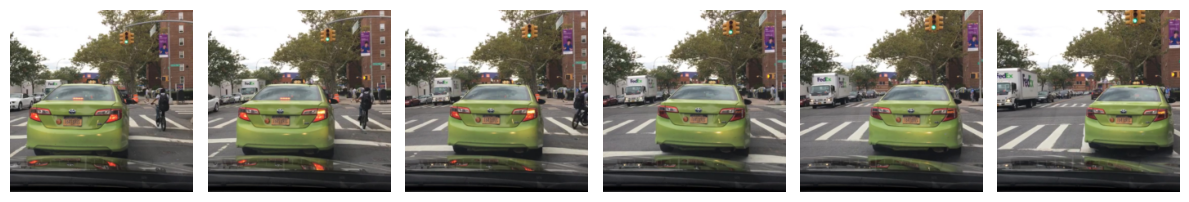}
        \caption{Unmasked sequence.}
    \end{subfigure}
    
    \vspace{0.5em}

    \begin{subfigure}[t]{0.49\linewidth}
        \centering
        \includegraphics[width=\linewidth]{figures/mask_generators/mask_generators_sobel_mask_overlay.png}
        \caption{Sobel gradient -- Mask Overlay}
        \label{fig:mask_gen_sobel_overlay}
    \end{subfigure}\hfill
    \begin{subfigure}[t]{0.49\linewidth}
        \centering
        \includegraphics[width=\linewidth]{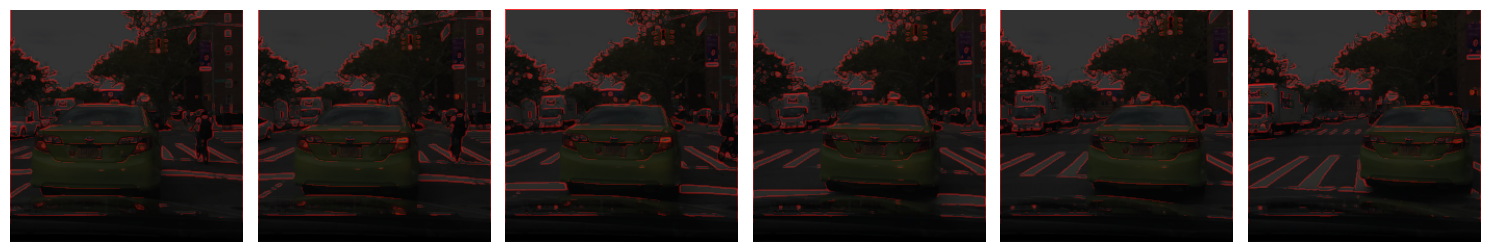}
        \caption{Sobel gradient -- Saliency Heatmap}
        \label{fig:mask_gen_sobel_heatmap}
    \end{subfigure}

    \vspace{0.5em}

    \begin{subfigure}[t]{0.49\linewidth}
        \centering
        \includegraphics[width=\linewidth]{figures/mask_generators/mask_generators_optical_flow_mask_overlay.png}
        \caption{Optical flow -- Mask Overlay}
        \label{fig:mask_gen_optical_flow_overlay}
    \end{subfigure}\hfill
    \begin{subfigure}[t]{0.49\linewidth}
        \centering
        \includegraphics[width=\linewidth]{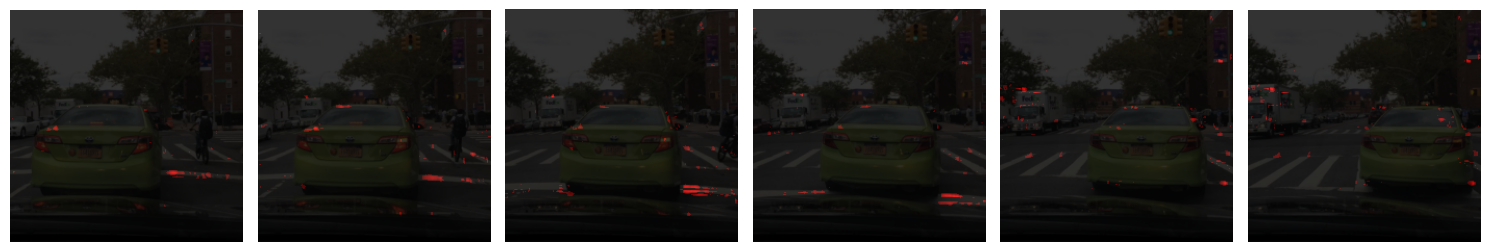}
        \caption{Optical flow -- Saliency Heatmap}
        \label{fig:mask_gen_optical_flow_heatmap}
    \end{subfigure}

    \vspace{0.5em}

    \begin{subfigure}[t]{0.49\linewidth}
        \centering
        \includegraphics[width=\linewidth]{figures/mask_generators/mask_generators_motion_comp_diff_mask_overlay.png}
        \caption{Motion-Compensated Differencing -- Mask Overlay}
        \label{fig:mask_gen_mcd_overlay}
    \end{subfigure}\hfill
    \begin{subfigure}[t]{0.49\linewidth}
        \centering
        \includegraphics[width=\linewidth]{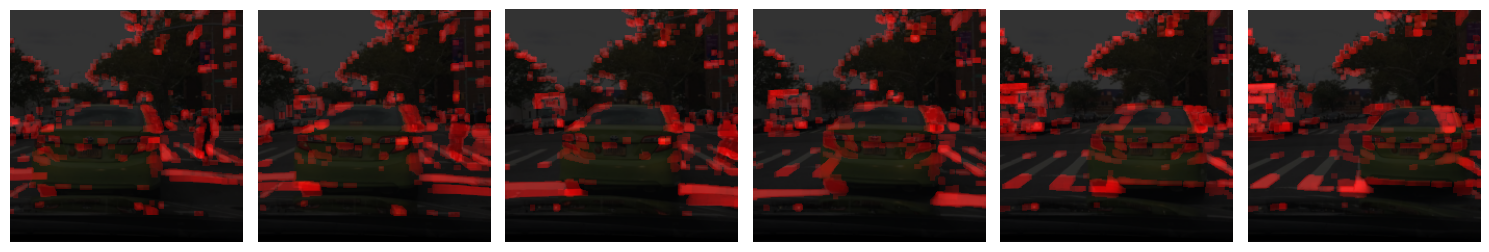}
        \caption{Motion-Compensated Differencing -- Saliency Heatmap}
        \label{fig:mask_gen_mcd_heatmap}
    \end{subfigure}

    \vspace{0.5em}

    \begin{subfigure}[t]{0.49\linewidth}
        \centering
        \includegraphics[width=\linewidth]{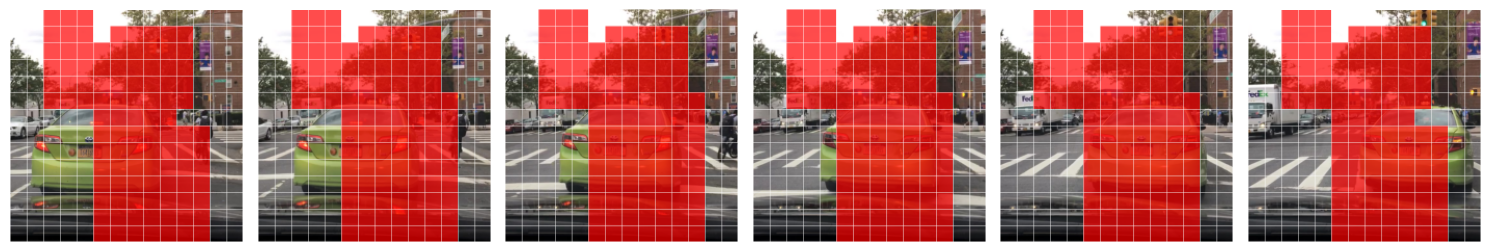}
        \caption{Combined Sobel + Flow -- Mask Overlay}
        \label{fig:mask_gen_combined_overlay}
    \end{subfigure}\hfill
    \begin{subfigure}[t]{0.49\linewidth}
        \centering
        \includegraphics[width=\linewidth]{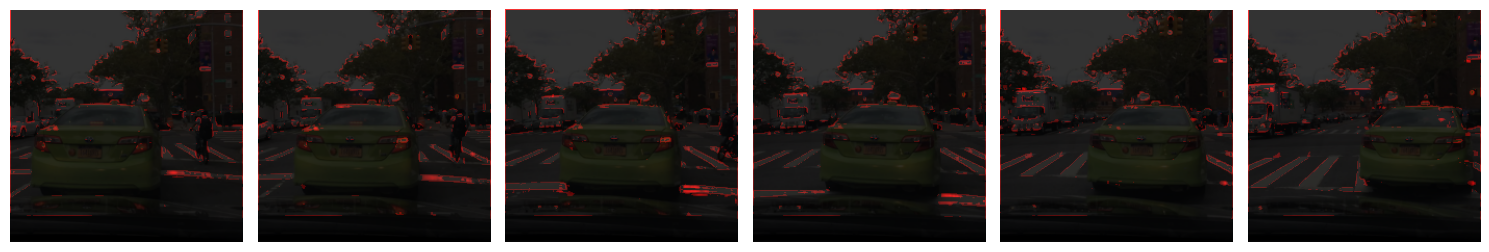}
        \caption{Combined Sobel + Flow -- Saliency Heatmap}
        \label{fig:mask_gen_combined_heatmap}
    \end{subfigure}

    \caption{Qualitative comparison of the four saliency-guided mask generators.
    Each row shows one generator: \textbf{(a--b)}~ , \textbf{(c--d)}~Optical flow, \textbf{(e--f)}~Motion-Compensated Differencing (\acrshort{mcd}), and \textbf{(g--h)}~Combined Sobel + Optical Flow.
    The left column displays the resulting mask overlay on the input frame, while the right column shows the corresponding saliency heatmap.}
    \label{fig:mask_generators_appendix}
\end{figure*}

\figref{fig:mask_generators_appendix} qualitatively compares the different saliency-based masking strategies.

The Sobel gradient primarily responds to high-frequency edge structures. However, it cannot distinguish between static texture and dynamic objects, as reflected by strong responses on lane markings, tree canopies, and other background edges.
In contrast, the optical-flow-based saliency map focuses on raw motion cues, but can be dominated by overall optical flow fields induced by egocentric camera motion.
Motion-compensated differencing instead compensates for camera ego-motion before computing frame differences. This results in stronger responses on motion boundaries, dynamic objects, and complex transition boundaries, such as the leading vehicle and current objects, while remaining robust to dominant uniform optical flow fields.
The combined Sobel and optical-flow strategy integrates complementary appearance and motion cues, but also inherits flow-related artifacts, for example along lane markings.

\section{Downstream Task Visualizations}
\label{app:downstream_demo}

\begin{figure*}[t]
    \centering

    \begin{subfigure}[t]{\linewidth}
        \centering
        \includegraphics[width=\linewidth]{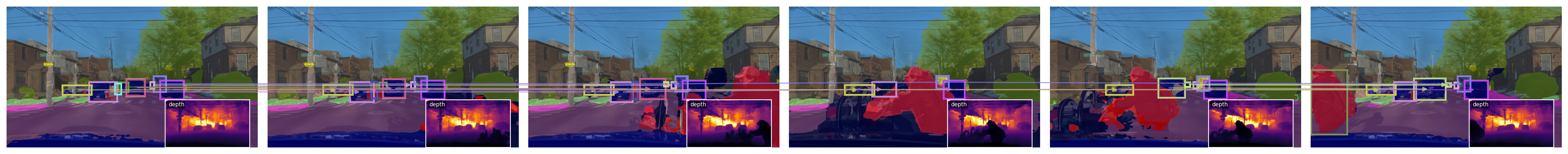}
        \caption{BDD100K b1d10d08-ec660956}
        \label{fig:downstream_appendix_bdd_b1d10d08}
    \end{subfigure}

    \vspace{0.5em}

    \begin{subfigure}[t]{\linewidth}
        \centering
        \includegraphics[width=\linewidth]{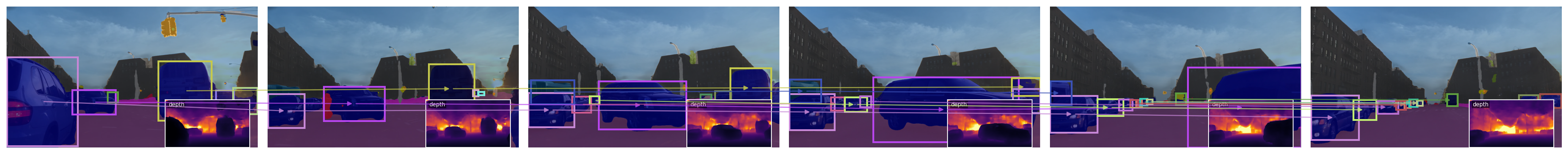}
        \caption{BDD100K b21c86ac-2eb7ba16}
        \label{fig:downstream_appendix_bdd_b21c86ac}
    \end{subfigure}

    \vspace{0.5em}

    \begin{subfigure}[t]{\linewidth}
        \centering
        \includegraphics[width=\linewidth]{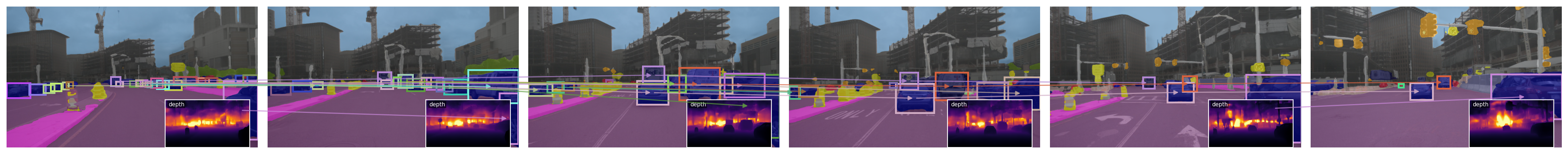}
        \caption{nuScenes scene-0253}
        \label{fig:downstream_appendix_scene0253}
    \end{subfigure}

    \vspace{0.5em}

    \begin{subfigure}[t]{\linewidth}
        \centering
        \includegraphics[width=\linewidth]{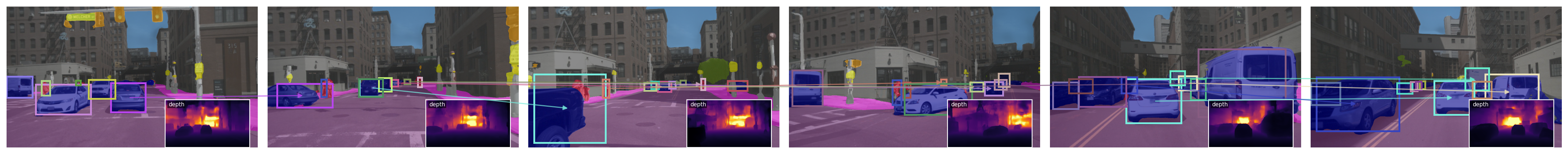}
        \caption{nuScenes scene-0397}
        \label{fig:downstream_appendix_scene0397}
    \end{subfigure}

    \vspace{0.5em}

    \begin{subfigure}[t]{\linewidth}
        \centering
        \includegraphics[width=\linewidth]{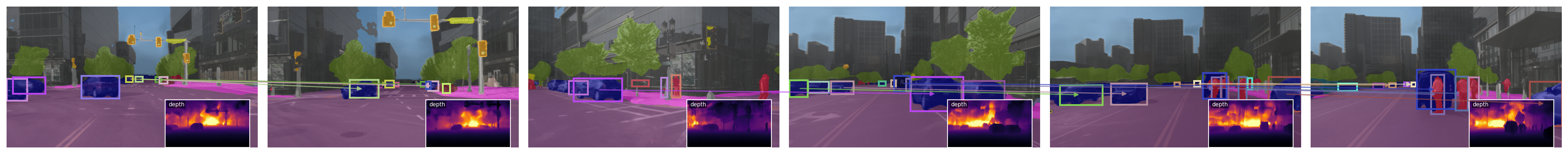}
        \caption{nuScenes scene-0518}
        \label{fig:downstream_appendix_scene0518}
    \end{subfigure}

    \vspace{0.5em}

    \caption{Additional downstream-task visualizations over six-frame sequences.
    The appendix figure collects the remaining BDD100K scenes b1d10d08-ec660956 and b21c86ac-2eb7ba16, together with nuScenes scenes 0253, 0397, and 0518.
    Each composite overlays semantic segmentation, monocular depth estimation, and multi-object tracking predictions.}
    \label{fig:downstream_demo_appendix}
\end{figure*}

\figref{fig:downstream_demo_appendix} presents qualitative results for downstream tasks over six-frame sequences.
Each visualization overlays semantic segmentation, monocular depth estimation, and multi-object tracking predictions on scenes from BDD100K (top) and nuScenes (bottom). 
These sequences are out of domain for the downstream models, which were fine-tuned on Cityscapes for semantic segmentation and KITTI-2015 for depth estimation.

Overall, the results demonstrate strong temporal consistency across all tasks. In particular, tracking remains robust throughout the sequences, including challenging scenarios such as turning maneuvers (e.g., \figref{fig:downstream_appendix_bdd_b21c86ac}) and interactions with deformable objects such as pedestrians (e.g., \figref{fig:downstream_appendix_scene0518}).
The dense semantic segmentation masks and depth prediction maps also show sharp object boundaries.

However, a few failure cases can be observed. 
Tracking failures occur after full occlusion, as illustrated by the SUV in the center of~\figref{fig:downstream_appendix_bdd_b1d10d08} and the white sedan in \figref{fig:downstream_appendix_scene0397}, which turns into the path of the ego-vehicle. 
The former case may be attributed to suboptimal hyperparameter choices in the ByteTrack tracker, while the latter reflects an inherently challenging traffic scene due to the rapid viewpoint and context change induced by the ego-vehicle's turn.
Additionally, segmentation artifacts are visible for partially occluded objects, such as the pedestrian in \figref{fig:downstream_appendix_bdd_b1d10d08}, where the predicted mask appears fragmented. 
Interestingly, this issue does not affect the corresponding depth estimate, which remains sharp and consistent. 
This discrepancy suggests that the artifacts likely originate from the downstream fine-tuning stage rather than from the underlying pre-trained features.

\section{Additional Downstream Task Experiments}
\label{app:additional_downstream}

\subsection{Extended Multi-Object Tracking Metrics}
\label{app:mot_extended}

The main paper reports only ID Switches (IDS) for the multi-object tracking evaluation, as it isolates re-identification quality 
when ground-truth detections are shared across all models.
In \tabref{tab:mot_extended} we report the compound MOT metrics, MOTA and HOTA, for completeness.
However, these metrics are less informative in our setting since they also incorporate detection performance, 
which is perturbed by our choice to rely on ground-truth detections for association of bounding boxes across frames.
The IDS column is reproduced from the main paper for reference.

\begin{table}[!htbp]
    \scriptsize
    \centering
    \caption{Extended multi-object tracking metrics on BDD100k~\cite{bdd100k}.
    All methods use ground-truth detection boxes to isolate re-identification quality, consistent with IDS metric in the main paper.
    MOTA and HOTA are compound metrics that additionally incorporate detection performance.}
    \label{tab:mot_extended}
    \begin{tabular}{@{}lcc ccc@{}}
        \toprule
        \textbf{Method} &
        \textbf{Model} &
        \textbf{MOTA $\uparrow$} &
        \textbf{HOTA $\uparrow$} &
        \textbf{IDS $\downarrow$} \\ \midrule
        DINOv2~\cite{dinov2}      & ViT-B & 68.99 & 78.96 & 10550 \\ \midrule
        V-JEPA4A (random)         & ViT-B & 71.96 & 80.86 & 10047 \\
        V-JEPA4A                  & ViT-B & 79.76 & 92.42 & \underline{9134}  \\
        V-JEPA4A (random, DINOv3) & ViT-L & 74.91 & 81.11 & 9906  \\
        V-JEPA4A (DINOv3 teacher) & ViT-L & 80.59 & 93.16 & \textbf{9018}  \\
        \bottomrule
    \end{tabular}
\end{table}

\subsection{Eval Benchmarks}
\label{app:full_metrics_late}

\tabref{tab:eval_benchmarks_supp} extends the evaluation benchmark table from the main paper 
by including an additional pre-trained encoder model that was trained with random multi-block masking and a DINOv3 frozen target encoder.
These results are a direct ablation of the \texttt{V-JEPA4A} model with \acrshort{mcd}-based masking and a DINOv3 frozen target encoder.
Replacing random masking with \acrshort{mcd}-based saliency masking under the same ViT-L/DINOv3 teacher setting yields gains across all three tasks: semantic segmentation improves from $51.9 \to 73.2$ mIoU, depth RMSE drops from $4.24$\,m to $3.75$\,m, and ID switches decrease from $9906$ to $9018$.
This consistent improvement confirms that the saliency-guided masking strategy adds value on top of the choice of a frozen teacher.

\begin{table}[!htbp]
    \scriptsize
    \centering
    \caption{Benchmark results for self-supervised pre-trained vision transformers. 
    For depth estimation and semantic segmentation, the pretrained models were integrated as frozen vision encoders in a \acrshort{dpt} model. 
    For multi-object tracking, the ground truth detection boxes were used to pool feature representations for similarity-based matching across frames. Therefore, we report $IDS$ to isolates association quality. 
    Semantic segmentation was evaluated on Cityscapes~\cite{cityscapes}, depth prediction on KITTI-2015~\cite{kitti}, and multi-object tracking on BDD100k~\cite{bdd100k}.
    All results were obtained on the respective validation sets.}
    \label{tab:eval_benchmarks_supp}
    \setlength{\tabcolsep}{3.5pt}
    \renewcommand{\arraystretch}{1.1}
    \begin{tabular}{@{}lcc cccc@{}}
        \toprule
        \textbf{Method} & 
        \textbf{Model} & 
        \shortstack[c]{\textbf{Patch}\\\textbf{size}} & 
        \shortstack[c]{\textbf{Input}\\\textbf{resolution}} & 
        \shortstack[c]{\textbf{Sem. seg.}\\\textbf{mIoU [\%] $\uparrow$}} & 
        \shortstack[c]{\textbf{Depth}\\\textbf{RMSE [m] $\downarrow$}} & 
        \shortstack[c]{\textbf{MOT}\\\textbf{IDS $\downarrow$}} \\ \midrule
        DINOv2~\cite{dinov2}      & ViT-L & $1{\times}14{\times}14$ & $224{\times}224$ & 69.3          & \textbf{3.57}    & 10283            \\
        DINOv3~\cite{dinov3}      & ViT-L & $1{\times}16{\times}16$ & $512{\times}512$ & \underline{72.0} & 4.10 & 9877             \\ \midrule
        V-JEPA4A (random)         & ViT-B & $1{\times}14{\times}14$ & $518{\times}518$ & 48.4          & 6.90             & 10047            \\
        V-JEPA4A                  & ViT-B & $1{\times}14{\times}14$ & $518{\times}518$ & 64.6          & 4.52             & \underline{9134} \\ \midrule
        V-JEPA4A (random, DINOv3) & ViT-L & $1{\times}16{\times}16$ & $512{\times}512$ & 51.9          & 4.24             & 9906             \\
        V-JEPA4A (DINOv3)         & ViT-L & $1{\times}16{\times}16$ & $512{\times}512$ & \textbf{73.2} & \underline{3.75} & \textbf{9018}    \\ \bottomrule
    \end{tabular}
\end{table}

\section{Vision Language Model Experiments}
\label{app:vlm_experiments}

This section describes the hyperparameter tuning for the VLM experiments (\secref{app:vlm_ablation}) and discusses qualitative examples for visual question answering and trajectory prediction (\secref{app:vlm_demo}).

\subsection{VLM Training Recipe Ablation}
\label{app:vlm_ablation}

\tabref{tab:vlm_ablation} ablates the \texttt{Qwen3-VJEPA4A} training recipe step by step.
Starting from a dataset-specific prompt format (DSF), we progressively introduce tag unification (TU), formatting tokens (FTok), fine-tuning with visual instruction tuning (VIT), and interleaved image/video training (IIV).
No single recipe dominates uniformly, but the interleaved image/video variant achieves the best aggregate performance across all three benchmarks, suggesting that mixed-modality instruction tuning improves generalization across scene-understanding tasks.
Tag unification standardizes the answer schema for our Qwen3-based models, enabling cross-benchmark evaluation of the same model; the main-paper baselines remain benchmark-specific reference points because their question formats and answer spaces are not directly interchangeable.
Transitioning from dataset-specific prompt formatting to a unified tag scheme yields a notable gain in binary question accuracy on NuScenes-MQA (0.54\,$\to$\,0.61), while introducing formatting tokens primarily benefits categorical recognition (0.64\,$\to$\,0.71).
On OmniDrive, the interleaved variant achieves the highest counterfactual-reasoning accuracy of all configurations (0.41).

\begin{table}[H]
    \scriptsize
    \centering
    \caption{Ablation of \texttt{Qwen3-VJEPA4A} training recipes across VLM driving benchmarks. 
    \textbf{Bold} = best, \underline{underline} = second best within each column.}
    \label{tab:vlm_ablation}
    {\setlength{\tabcolsep}{2.5pt}
    \begin{tabular}{@{}l *{10}{c}@{}}
        \toprule
        \textbf{Recipe} & \multicolumn{3}{c}{\textbf{NuScenes}} & \multicolumn{3}{c}{\textbf{OmniDrive}} & \multicolumn{4}{c}{\textbf{Impromptu VLA}} \\
        \cmidrule(lr){2-4} \cmidrule(lr){5-7} \cmidrule(lr){8-11}
        & {\shortstack{Y/N \\ Acc.}} & {\shortstack{Cat. \\ Acc.}} & {\shortstack{Cat./Cnt. \\ L1}} & {\shortstack{Drive \\ dec.}} & {\shortstack{Act. \\ plan.}} & {\shortstack{Counterf. \\ reas.}} & {\shortstack{VRU det. \\ Acc.}} & {\shortstack{Behav. \\ justif.}} & {\shortstack{Act. plan. \\ Acc.}} & {\shortstack{Waypoint \\ L1}} \\
        \midrule
        DSF            & 0.54 & 0.62 & 0.34 & 0.15 & 0.30 & 0.15 & 0.18 & 0.55 & 0.30 & 0.80 \\
        + TU           & \textbf{0.61} & 0.64 & 0.34 & 0.21 & 0.33 & 0.19 & 0.23 & 0.59 & 0.32 & 0.60 \\
        + FTok         & 0.59 & 0.71 & 0.33 & 0.20 & \textbf{0.36} & 0.36 & 0.22 & 0.61 & 0.58 & 0.58 \\
        + VIT          & 0.60 & \textbf{0.73} & 0.31 & \textbf{0.22} & 0.34 & 0.38 & 0.24 & 0.60 & 0.54 & 0.62 \\
        + IIV          & \textbf{0.62} & 0.72 & \textbf{0.38} & 0.16 & 0.35 & \textbf{0.41} & \textbf{0.25} & \textbf{0.61} & \textbf{0.59} & 0.39 \\
        \bottomrule
    \end{tabular} 
    }

    \vspace{2pt}
    \parbox{\linewidth}{\scriptsize\textit{Recipe:} DSF = dataset-specific format; TU = tag unification; FTok = formatting tokens; VIT = fine-tuning with visual instruction tuning; IIV = interleaved image/video fine-tuning.}
\end{table}

\subsection{Qualitative Driving VLM Example}
\label{app:vlm_demo}

\begin{figure}[ht]
    \centering
    \includegraphics[width=0.32\linewidth]{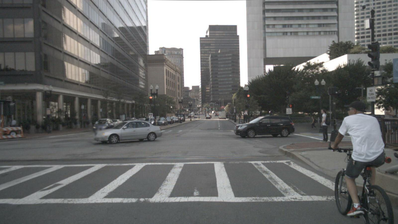}\hfill%
    \includegraphics[width=0.32\linewidth]{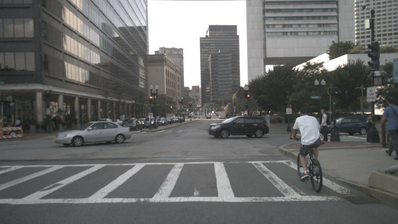}\hfill%
    \includegraphics[width=0.32\linewidth]{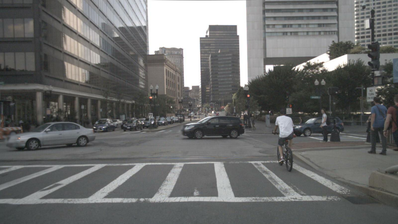}

    \vspace{0.5em}
    \begin{tcolorbox}[
        colback=gray!12, colframe=gray!50, arc=5pt, boxrule=0.6pt,
        left=5pt, right=5pt, top=3pt, bottom=3pt,
        fontupper=\footnotesize
    ]
        \textbf{Q:}\enspace Are there any traffic elements that could influence your driving decisions?
    \end{tcolorbox}

    \vspace{0.3em}
    \begin{minipage}[t]{0.485\linewidth}
        \begin{tcolorbox}[
            colback=Salmon!12, colframe=Salmon!55, colbacktitle=Salmon!40,
            coltitle=black, arc=5pt, boxrule=0.6pt,
            left=4pt, right=4pt, top=2pt, bottom=2pt,
            title={\scriptsize\bfseries Baseline (random masking)},
            fontupper=\scriptsize
        ]
            Yes, there are traffic elements: a black SUV is approaching the intersection, a cyclist is riding in the intersection, and traffic lights are visible.
        \end{tcolorbox}
    \end{minipage}\hfill
    \begin{minipage}[t]{0.485\linewidth}
        \begin{tcolorbox}[
            colback=LimeGreen!10, colframe=OliveGreen!55, colbacktitle=OliveGreen!30,
            coltitle=black, arc=5pt, boxrule=0.6pt,
            left=4pt, right=4pt, top=2pt, bottom=2pt,
            title={\scriptsize\bfseries \texttt{V-JEPA4A} VLM (ours)},
            fontupper=\scriptsize
        ]
            The traffic light ahead is \textbf{red}, requiring me to stop. There are multiple \textbf{pedestrians} near the crosswalks ahead and to the left. I need to watch out for the \textbf{cyclist in my lane}. Cars are crossing the intersection in front of us.
        \end{tcolorbox}
    \end{minipage}

    \caption{Qualitative comparison of a baseline VLM (random-masking backbone) and our saliency-guided \texttt{V-JEPA4A} backbone on a representative driving scene. Saliency-based pre-training produces more actionable, safety-critical scene descriptions: \texttt{V-JEPA4A} explicitly identifies the \emph{red} traffic light, pedestrians at the crosswalk, and the cyclist in the ego lane, whereas the baseline gives a vague enumeration that omits traffic-light state and pedestrian right-of-way.}
    \label{fig:demo_figure}
\end{figure}

\figref{fig:demo_figure} compares a baseline VLM with a random-masking-based backbone against \texttt{V-JEPA4A} with a saliency-guided backbone on a representative driving scene.
The example suggests that saliency-guided pre-training encourages attention to dynamic, safety-critical regions: \texttt{V-JEPA4A} correctly identifies the \emph{red} traffic light, pedestrians crossing ahead, and a cyclist in the ego lane, and describes them in actionable terms missing from the baseline output.

\begin{figure}[ht]
    \centering
    \begin{subfigure}{0.49\linewidth}
        \includegraphics[width=\linewidth]{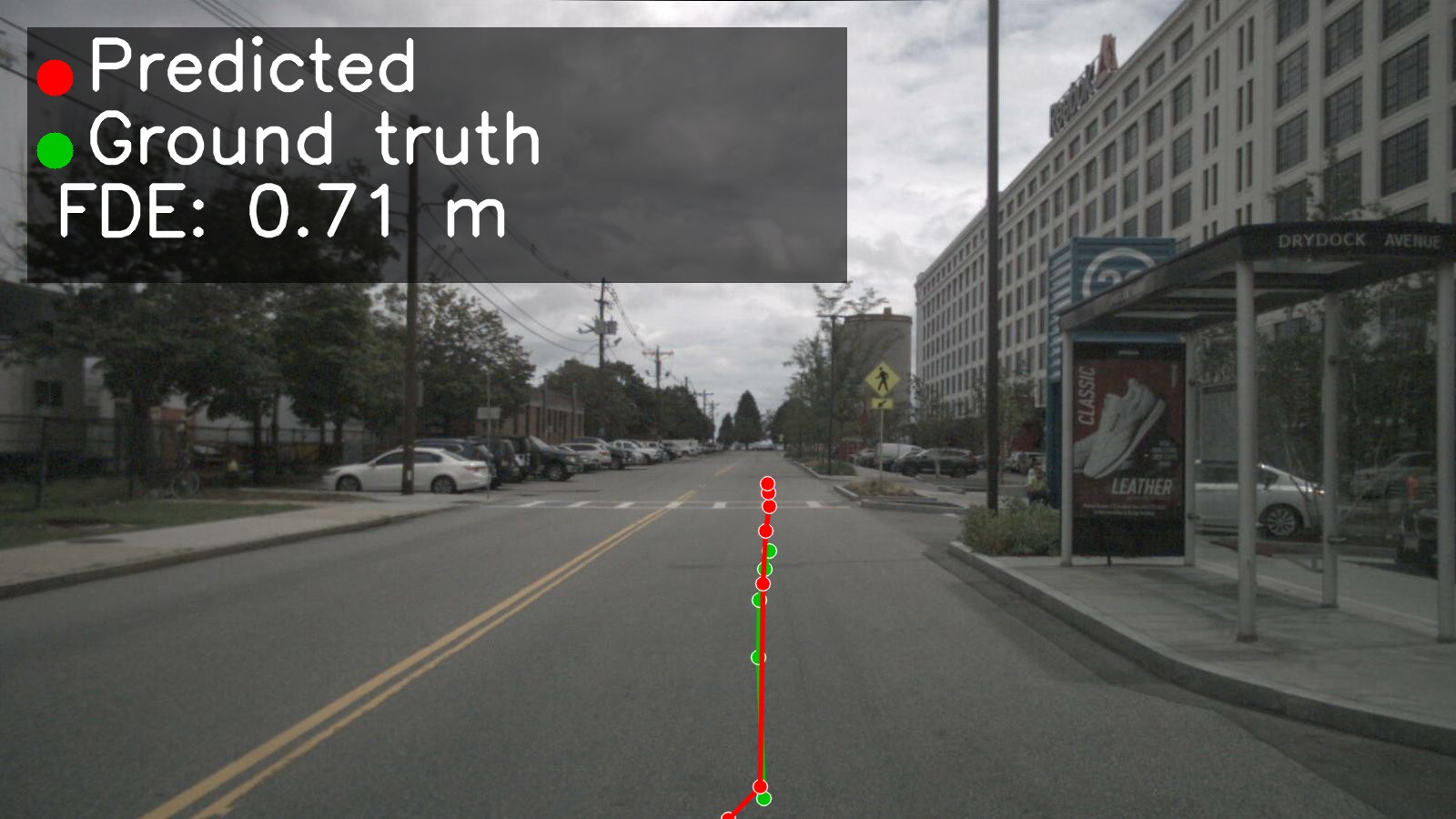}
        \caption{\textbf{Accurate prediction} (FDE\,=\,0.71\,m). Clear lane markings provide sufficient road structure for the model to closely follow the intended trajectory.}
        \label{fig:traj_good}
    \end{subfigure}\hfill
    \begin{subfigure}{0.49\linewidth}
        \includegraphics[width=\linewidth]{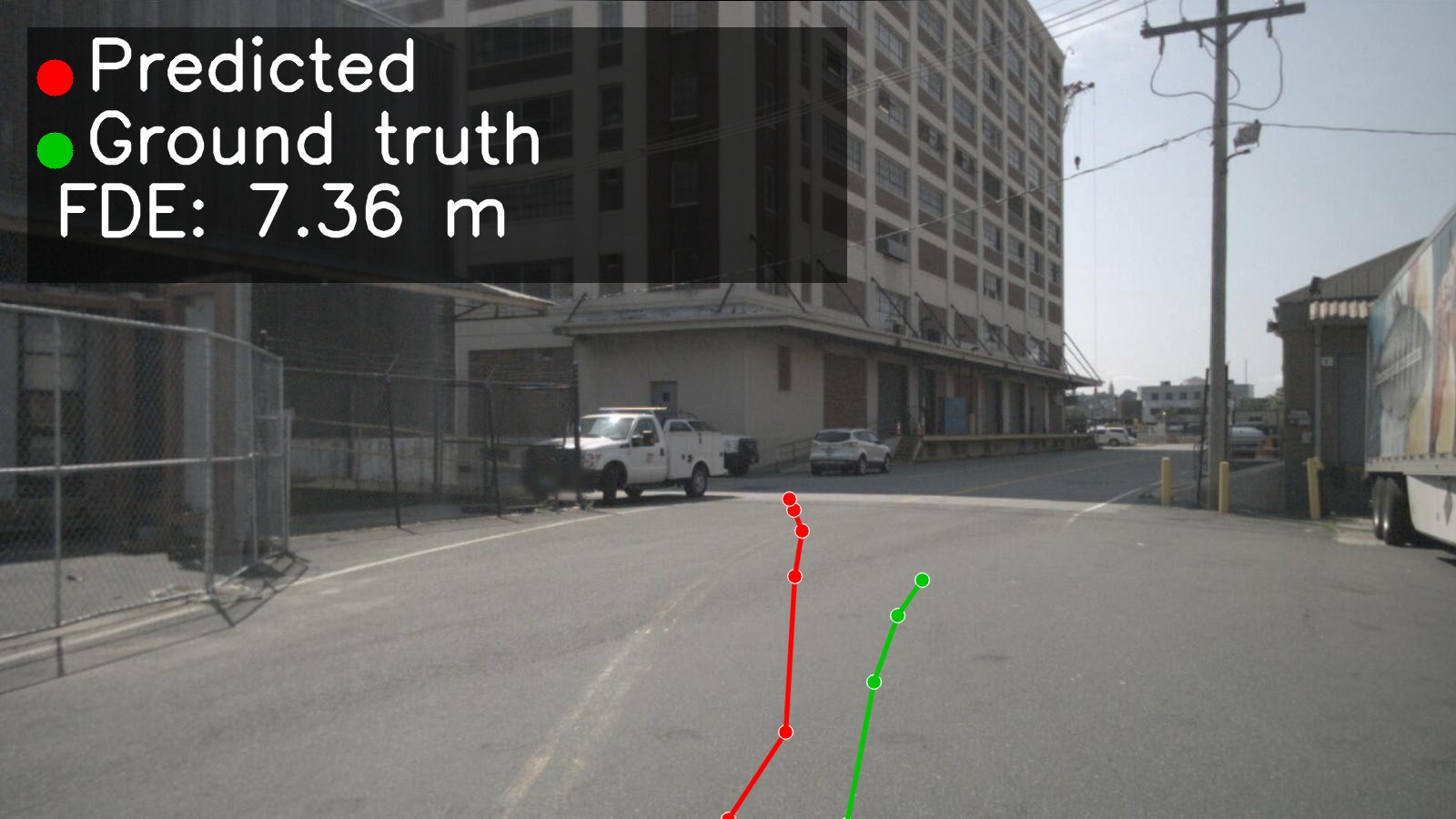}
        \caption{\textbf{Failure case} (FDE\,=\,7.36\,m). Washed-out lane markings leave the model without reliable lane cues, causing a large deviation from the ground-truth path.}
        \label{fig:traj_hard}
    \end{subfigure}
    \caption{Qualitative trajectory prediction examples from \texttt{Qwen3-VJEPA4A} on NuScenes.
    Predicted trajectories are shown in \textcolor{red}{red}; ground-truth future paths in \textcolor{green}{green}.}
    \label{fig:traj_examples}
\end{figure}

\figref{fig:traj_examples} presents two contrasting open-loop trajectory predictions from \texttt{Qwen3-VJEPA4A} on NuScenes.
These examples indicate that prediction quality is closely tied to the visual contrast of lane markings: clear road structure (\figref{fig:traj_good}, FDE\,=\,0.71\,m) enables the model to follow the intended path, whereas low-contrast, washed-out markings (\figref{fig:traj_hard}, FDE\,=\,7.36\,m) provide unreliable lane cues and lead to large lateral deviations. 

\end{document}